\documentclass[10pt,journal,compsoc]{IEEEtran}
\usepackage{times}
\usepackage{color,soul}
\usepackage{array}
\usepackage{amsmath,amssymb,amsfonts}
\usepackage{graphicx,subfigure,epstopdf,epsfig}
\usepackage{multirow,booktabs}
\usepackage[
            pdfstartview=FitH,
            CJKbookmarks=true,
            bookmarksnumbered=true,
            bookmarksopen=true,
            colorlinks, 
            linkcolor=red,
            anchorcolor=green,
            citecolor=blue
            ]{hyperref}
\usepackage{bookmark}
\usepackage{algorithm}
\usepackage{algorithmic}
\usepackage{yhmath}
\graphicspath{{/}}

\renewcommand\thetable{\Roman{table}}

\newcommand{\eref}[1]
{(\ref{#1})}
\newcommand{\fref}[1]
{\mbox{Fig.~\ref{#1}}}
\newcommand{\tref}[1]
{\mbox{Tab.~\ref{#1}}}

\begin{document}
%
\title{A Fast Ellipse Detector Using Projective Invariant Pruning}

\author{Qi~Jia,
        Xin~Fan,~\IEEEmembership{Member,~IEEE,}
        Zhongxuan~Luo,
        Lianbo~Song,
        and~Tie~Qiu
\IEEEcompsocitemizethanks{\IEEEcompsocthanksitem Q. Jia, X. Fan, Z. Luo, L. Song, and T. Qiu are with the School
of Software, Dalian University of Technology, China, and also with the Key Laboratory for Ubiquitous Network and Service Software of Liaoning Province.\protect\\
E-mail: xin.fan@ieee.org
\IEEEcompsocthanksitem Z. Luo is also with the School of Mathematical Sciences, Dalian University of Technology, China.
\IEEEcompsocthanksitem This work was supported in part by the National Natural Science Foundation of China under Grant 61402077, Grant 61432003, Grant 61328206, Grant 11171052.
}}

\markboth{}
{Shell \MakeLowercase{\textit{et al.}}: Bare Demo of IEEEtran.cls for Journals}

\IEEEtitleabstractindextext{%
\begin{abstract}
Detecting elliptical objects from an image is a central task in robot navigation and industrial diagnosis where the detection time is always a critical issue. Existing methods are hardly applicable to these real-time scenarios of limited hardware resource due to the huge number of fragment candidates (edges or arcs) for fitting ellipse equations. In this paper, we present a fast algorithm detecting ellipses with high accuracy. The algorithm leverage a newly developed projective invariant to significantly prune the undesired candidates and to pick out elliptical ones. The invariant is able to reflect the intrinsic geometry of a planar curve, giving the value of $-1$ on any three collinear points and $+1$ for any six points on an ellipse. Thus, we apply the pruning and picking by simply comparing these binary values. Moreover, the calculation of the invariant only involves the determinant of a $3\times3$ matrix. Extensive experiments on three challenging data sets with 650 images demonstrate that our detector runs 20\%-50\% faster than the state-of-the-art algorithms with the comparable or higher precision.
\end{abstract}

}

\maketitle

\IEEEdisplaynontitleabstractindextext

\IEEEpeerreviewmaketitle

\IEEEraisesectionheading{\section{Introduction}\label{sec:introduction}}

\IEEEPARstart{E}{llipses} are quite common in natural or artificial scenes. The detection of ellipses in a fast and reliable manner from real world images provides a powerful analysis tool for many computer vision applications such as wheels detection~\cite{cooke2010fast}, biological cell division~\cite{zafari2015segmentation}, and object segmentation for industrial applications~\cite{teutsch2006real}. Ellipse detection still remains unresolved as one of the classical tasks with long history. Most existing methods perform far from real time, which hinder their applications in reality.

The earliest ellipse detection algorithm dates back to the classical Hough transform (HT) that fits the parametric form of an ellipse using a voting scheme~\cite{Illingworth1988}. The standard HT approach extracts ellipses by finding the clusters in a five-dimensional (5D) parametric space, consuming a great deal of memory and time. The randomized HT (RHT) improves the performance by reducing the number of false alarms~\cite{mclaughlin1998randomized}. The iterative RHT (IRHT) speeds up RHT so significant by focusing on the candidates likely to be an ellipse that it only needs 1-D accumulators~\cite{lu2008detection}. However, both RHT and IRHT are still quite slow attributing to the voting processing among numerous candidates, and the geometry relationships between points are also neglected during voting.

Researchers introduce algebraic or geometric constraints on \emph{points} of an ellipse to screening candidates. Liang~\emph{et~al.}~\cite{liang2015robust} introduce the maximum correntropy criterion into the constrained least-square fitting to alleviate the influence of outliers. Mulleti~\emph{et~al.}~\cite{mulletiellipse} use the finite rate of innovation sampling principle to fit noisy or partial ellipse. Both methods produce ellipses with less bias. However, they can only work on the image with one single ellipse.
Xie and Ji exploit the symmetry of two points on ellipse, reducing the voting parameter to one~\cite{xie2002new}. However, it is time-prohibitive to enumerate every point pairs as elliptical candidates. Basca ~\emph{et~al.}~\cite{bascca2005randomized} accelerate Xie~\emph{et~al.}'s method with RHT by considering only a small random subset of initial point pairs. Zhang and Liu~\cite{zhang2005robust} use edge directional properties to reduce point combinations that lie out of the same ellipse boundary.

Many methods take into account the geometric constraints on \emph{arc} segments as the symmetry between points brings too many candidate pairs. Kim~\emph{et~al.}~\cite{kim2002fast} extract arc segments approximated by short straight lines. Libuda~\emph{et~al.}~\cite{libuda2007ellipse} and Prasad~\emph{et~al.}~\cite{prasad2013ellifit} improve Kim \emph{et al.}'s method with less memory usage. However, these candidate arcs connecting short line segments may merge intersected arcs from different ellipses, resulting in lower precisions~\cite{kim2002fast,libuda2007ellipse}. Nguyen~\emph{et~al.}~\cite{nguyen2009real} detect ellipses upon arcs by edge grouping. Their method is able to handle incomplete ellipses, but fails to detect ellipses splitting into many short arcs. Some other works formulate the mergence of elliptical arcs as an assignment problem, and iteratively correct the detection results~\cite{mai2008hierarchical,chia2011split,lu2015effective}. These methods have high detection rates, but suffer from heavy computational costs. Prasad~\emph{et~al.}~\cite{prasad2010clustering} merge elliptical arcs with the relationship score given by the center of the ellipse fitting the arcs. Recently, Fornaciari~\emph{et~al.}~\cite{fornaciari2014fast} develop an ellipse detector that classifies elliptical arcs into four groups and estimates the ellipse parameters using the decomposed parameter space. There are still a number of candidate arcs in each group while their method renders a relatively faster detection than previous methods. Especially, it is quite time consuming to calculate every possible combinations of arcs from each group, not to mention that many of them are wrong combinations.

All aforementioned methods start the estimation from points or arcs with their positional constraints. Actually, there exist more constraints whether points or arcs are amenable to the analytical ellipse equation. In~\cite{sugaya2010ellipse}, RANSAC is used to randomly choose five points repeatedly until the ellipse determined by these five points closely passes through a maximum number of edge points. This method is inefficient as there exist so many combinations of five points and one has to calculate ellipse parameters for each five-point combination. Instead of directly calculating the ellipse parameters, Liu and Hu~\cite{liu2013effective} use geometric distances of points to a conic in order to evaluate the similarity between any two of selected arcs. There are still a large number of wrong arcs combinations, and the computation of distances between points and arcs lower down its efficiency.

In this paper, we circumvent the high computational load by pruning and picking candidates using a projective invariant, named characteristic number (CN)~\cite{luo2014projective}. The projective invariance of CN is introduced in~\cite{fan2015fiducial} acting as geometric constraints for fiducial point localization under face pose changes. Later, Jia~\emph{et~al.} employ this invariant property to construct a shape descriptor robust to perspective deformations~\cite{jia2016hierarchical}. For the first time, we explicitly take the advantage of the CN property giving the characterization of the intrinsic geometry of an underlying planar curve of points. The invariant gives the value of $-1$ on any three collinear points and $+1$ for any six points on a conic (ellipse or other conic curves). We are able to efficiently pruning straight lines and to determine whether non-collinear six points belong to the same conic without need of analytically fitting nor voting. The calculation of CN is also quite fast since the determinants of $3\times3$ matrices constitute the primitives of CN.

\begin{figure*}
    \centering
    \subfigure[] { \includegraphics[width=1.6in]{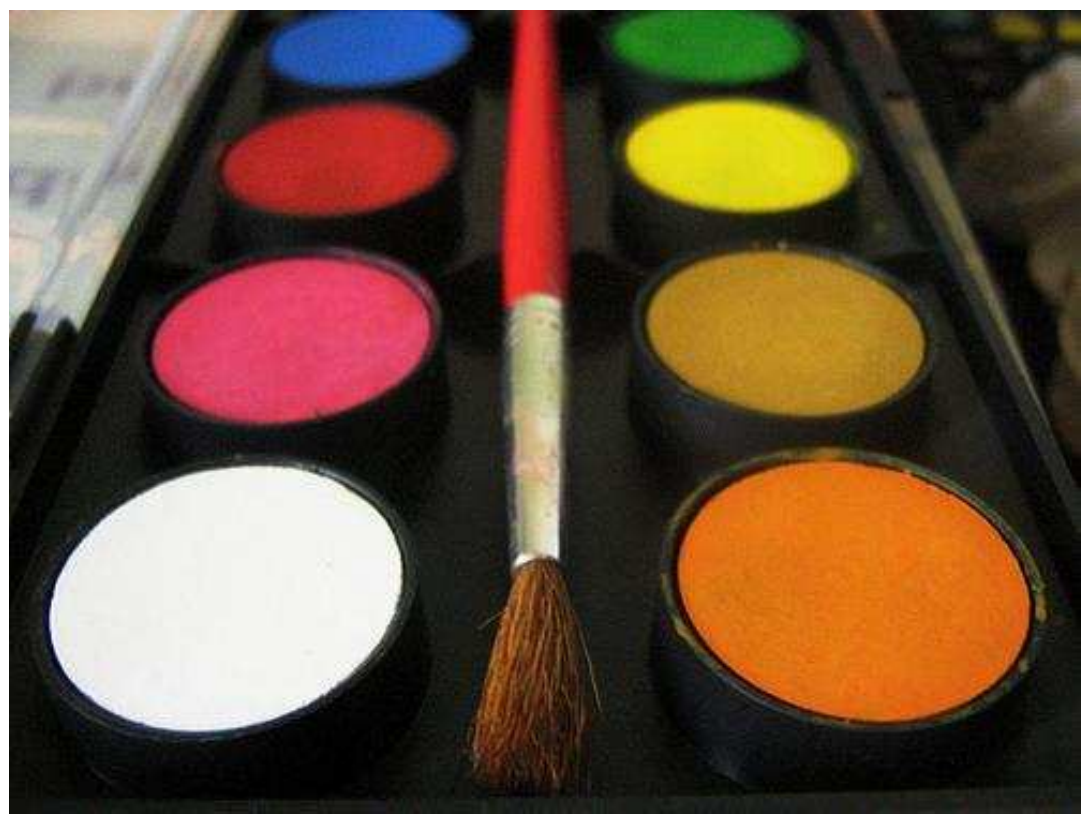} \label{fig:process1} }
    \subfigure[] { \includegraphics[width=1.6in]{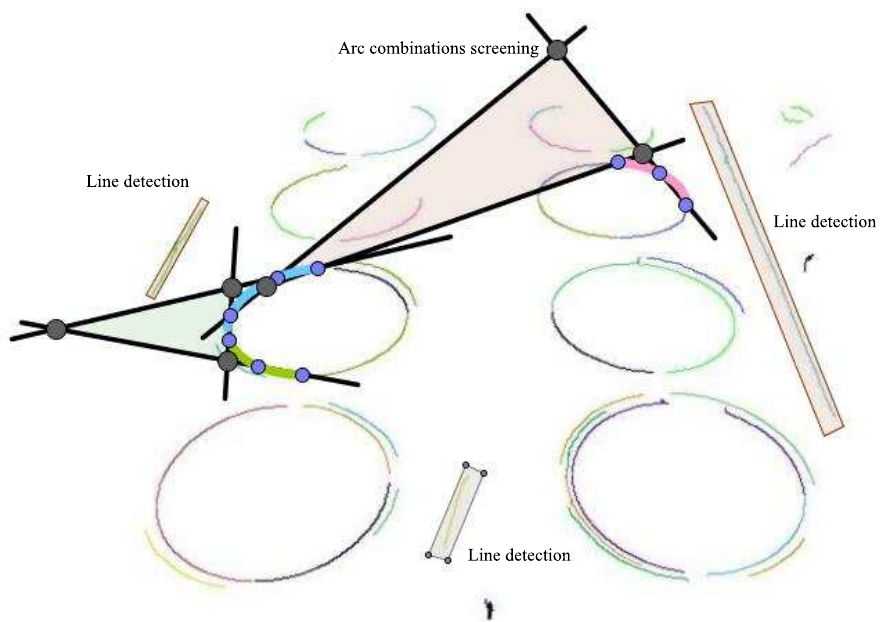} \label{fig:process2} }
    \subfigure[] { \includegraphics[width=1.6in]{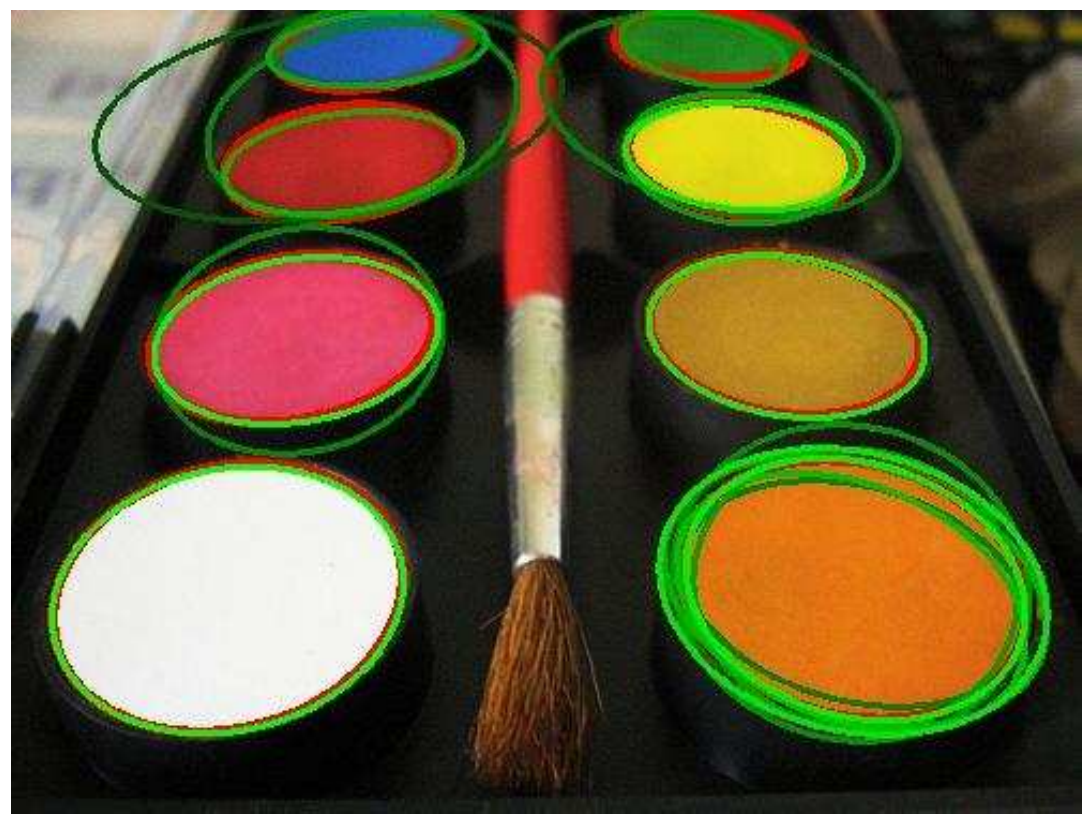} \label{fig:process3} }
    \subfigure[] { \includegraphics[width=1.6in]{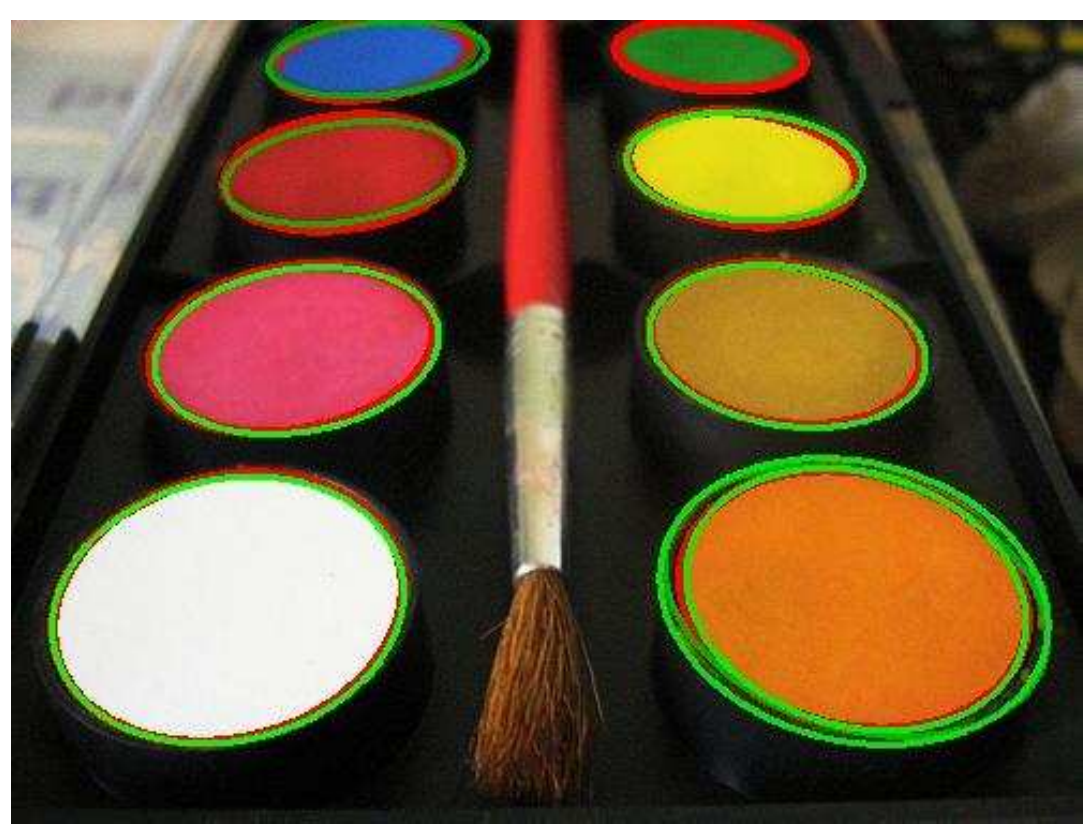} \label{fig:process4} }
    \caption{Ellipse detection process: (a) input image, (b) line pruning and arc picking, (c) candidate ellipse given by picked arcs, and (d) final detection with parameter fitting and ellipse validation.} \label{fig:process}
\end{figure*}

Figure~\ref{fig:process2} illustrates the pruning and picking on an input image  with a bunches of ellipses in~\fref{fig:process1}. Candidate arc segments appear different colors, and we are able to prune those line segments having the CN value $-1$. We use six points on two arc segments, three on each, to calculate a CN value. The points on the bold arcs in pink and blue form the light brown triangle, while those on blue and green bold arcs generate the light green triangle. These triangles (closed loops) formed by auxiliary lines connecting arc points are necessary for CN calculation. The six points on the arc pairs from the same ellipse, e.g., the blue and green pair in~\fref{fig:process2}, give the CN value $+1$. Thus, the arcs belonging to one ellipse easily pop out by CN values, and~\fref{fig:process3} shows the ellipse candidates formed by these picked arc segments. Figure~\ref{fig:process4} presents the final detection results by further ellipse fitting and validation. The proposed method is robust with high efficiency, suitable for real-time applications as validated in experiments.

The rest of this paper is organized as follows. Section 2 introduces the characteristic number (CN) and its property on lines and conic curves. Section 3 elaborates our fast ellipse detection algorithm with line pruning and arc picking using CN. Section 4 demonstrates experimental evaluations on accuracy and efficiency. Section 5 concludes the paper.

\section{Characteristic number on line and conic}
In this section, we first introduce the general definition of the characteristic number (CN), and give its computation on three collinear points (CNL) and six points in a conic curve (CNC), respectively. Also, we present the CN properties for these two configurations, the key to efficiently pruning line segments and picking candidates for elliptical arcs.
\subsection{Characteristic number}
The characteristic number extends the classical cross ratio in various respects, and reflects the intrinsic geometry underlying given points. The CN value of three collinear points is $-1$, while six points lying on a conic curve have a common CN value $+1$. We give the definition of $CN$ below~\cite{luo2014projective}.

\textbf{Definition 1}
Let $\mathbb{P}^{m}$ be $m$-dimension projective space over $\mathbb{K}$, and $P_1,P_2,\ldots, P_r$ be $r$ distinct points in $\mathbb{P}^{m}$ that construct a closed loop ($P_{r+1}=P_1$). There are $n$ distinct points $Q_i^{(1)},\ldots Q_i^{(n)}$ on the line segment $P_iP_{i+1}, i = 1,\ldots,r.$ Each point $Q_i^{(j)}$ can be linearly represented by $P_i$ and $P_{i+1}$ as
\begin{equation}
Q_i^{(j)} = a_i^{(j)}{P_i} + b_i^{(j)}{P_{i + 1}}.
\end{equation}
Let ${\cal P} = \{ {P_i}\} _{i = 1}^r$ and ${\cal Q} {\rm{ =  \{ }}Q_i^{(j)}\} _{i = 1,...,r}^{j = 1,...,n}$, the quantity
\begin{equation}\label{eq:CN}
CN({\cal P},{\cal Q}{\rm{) = }}\prod\limits_{i = 1}^r {\left( {\prod\limits_{j = 1}^n {\frac{{b_i^{(j)}}}{{a_i^{(j)}}}} } \right)},
\end{equation}
is defined as the \textbf{characteristic number} of ${\cal P}$ and ${\cal Q}$.

The classical cross ratio is a special case of CN when $r=2$ and $n=2$. The characteristic number is proven to be a projective invariant, meaning that CN values keep unchanged under projective/perspective transformations.

The definition of CN requires a closed loop with $r$ sides, and the number of points $Q^{(j)}$ at every side has to be identical. Subsequently, we draw auxiliary lines to set up this geometric configuration required for calculating CN on three collinear points and six points on a conic curve (ellipse).

\subsection{Characteristic number on three collinear points}
Three pairwise intersecting lines $a$, $b$ and $c$ intersect at three points $P_1$, $P_2$ and $P_3$, forming a closed loop. These three lines intersect another line $l$ at three points $Q_1^{(1)}$, $Q_2^{(1)}$ and $Q_3^{(1)}$, as shown in~\fref{fig:threeCN}. According to Definition 1, $Q_1^{(1)}$, $Q_2^{(1)}$ and $Q_3^{(1)}$ can be linearly represented by $P_1$, $P_2$ and $P_3$ as:

\begin{figure}
\centering
\includegraphics[height=6.5cm]{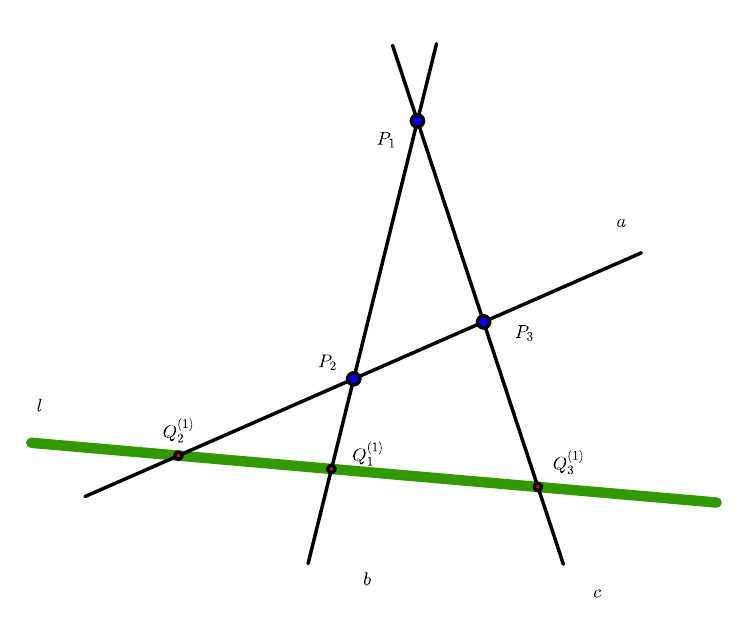}
\caption{Characteristic number on three collinear points (CNL), $Q_1^{(1)}$, $Q_2^{(1)}$ and $Q_3^{(1)}$. Three pairwise intersecting lines $a$, $b$ and $c$ form a closed loop $\triangle P_1P_2P_3$. }
\label{fig:threeCN}
\end{figure}
\begin{equation}
\label{eq:CNL_coef}
\left\{ \begin{gathered}
  {Q_1}^{(1)} = {a_1}^{(1)}P_1 + {b_1}^{(1)}P_2, \hfill \\
  {Q_2}^{(1)} = {a_2}^{(1)}P_2 + {b_2}^{(1)}P_3, \hfill \\
  {Q_3}^{(1)} = {a_3}^{(1)}P_3 + {b_3}^{(1)}P_1, \hfill \\
\end{gathered}  \right.
\end{equation}

Substituting the coefficients into~\eref{eq:CN}, we calculate CN on three collinear points (CNL) as~\eref{eq:CNL}. It can be proved that the CN value equals to $-1$ if $Q_1^{(1)}$, $Q_2^{(1)}$ and $Q_3^{(1)}$ are collinear~\cite{fan2015fiducial}.
\begin{equation}\label{eq:CNL}
CNL({\cal P},{\cal Q}) = CN({\cal P},{\cal Q}) = \prod\limits_{i = 1}^3 {\frac{{{b_i}^{(1)}}}{{{a_i}^{(1)}}}}  =  - 1,
\end{equation}
The proof process also implies that the calculation of CNL is achieved by the determinant of a $3 \times 3$ matrix. In this study, we use this property to prune line segments in ellipse detection.

\subsection{Characteristic number on six points of a conic}
\begin{figure}
\centering
    \includegraphics[width=0.45\textwidth]{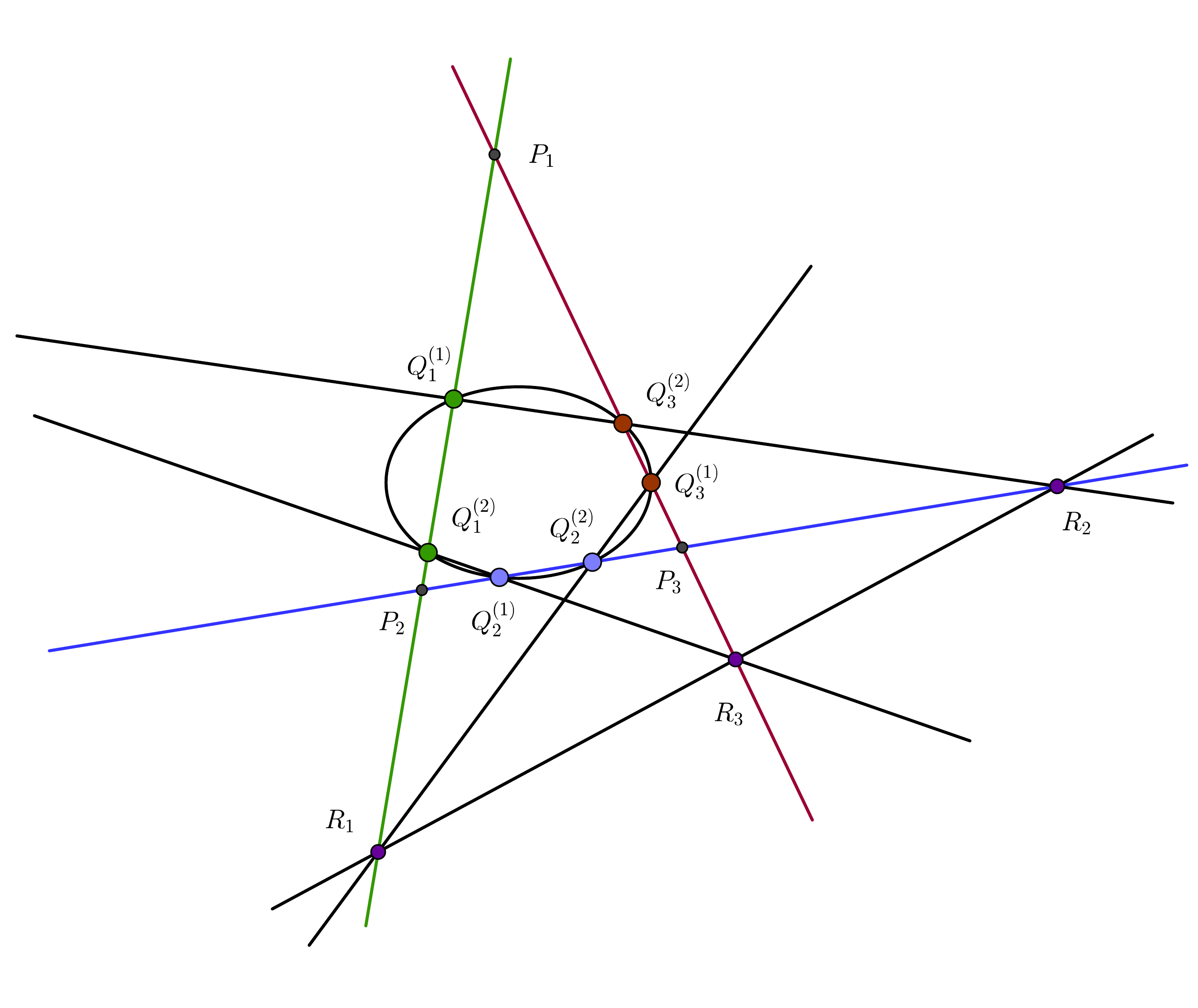}
    \caption{Characteristic number on six points of a conic (CNC). The closed loop $\triangle P_1P_2P_3$ intersects a conic (ellipse) at two points on each side as $\{{Q_i}^{(j)}|i=1,2,3;j=1,2\}$. The points $R_i(i=1,2,3)$ are auxiliary points to prove $CNC=+1$. }
    \label{fig:ProveCNC}
\end{figure}

We denote the line through two points ${{Q_1}^{(1)}}$ and ${{Q_1}^{(2)}}$ as ${{Q_1}^{(1)}}{{Q_1}^{(2)}}$, and the intersection of two lines as $<{{Q_1}^{(1)}}{{Q_1}^{(2)}}, {{Q_2}^{(1)}}{{Q_2}^{(2)}}>$. Let $\{{Q_i}^{(j)}|i=1,2,3;j=1,2\}$ be six distinct points on a conic (ellipse) as shown in~\fref{fig:ProveCNC}, and $P_1$, $P_2$ and $P_3$ be three intersection points of the lines connecting some pairs of the six points on the conic:

\begin{equation}\label{eq:p13q16}
\begin{array}{*{20}{c}}
  {{P_1} =  < {Q_3}^{(1)}{Q_3}^{(2)} ,{Q_1}^{(1)}{Q_1}^{(2)}  > ,} \\
  {{P_2} =  < {Q_1}^{(1)}{Q_1}^{(2)} ,{Q_2}^{(1)}{Q_2}^{(2)}  > ,} \\
  {{P_3} =  < {Q_2}^{(1)}{Q_2}^{(2)} ,{Q_3}^{(1)}{Q_3}^{(2)}  > .}
\end{array}
\end{equation}
Similar to~\eref{eq:CNL_coef}, each point of $\{{Q_i}^{(j)}|i=1,2,3;j=1,2\}$ can be linearly represented by a pair of points from $\{P_1,P_2,P_3\}$.
\begin{equation}\label{eq:p}
\begin{array}{*{20}{c}}
  {{Q_1}^{(1)} = {a_1}^{(1)}P_1 + {b_1}^{(1)}P_2,} \\
  {{Q_1}^{(2)} = {a_1}^{(2)}P_1 + {b_1}^{(2)}P_2,} \\
  {{Q_2}^{(1)} = {a_2}^{(1)}P_2 + {b_2}^{(1)}P_3,} \\
  {{Q_2}^{(2)} = {a_2}^{(2)}P_2 + {b_2}^{(2)}P_3,} \\
  {{Q_3}^{(1)} = {a_3}^{(1)}P_3 + {b_3}^{(1)}P_1,} \\
  {{Q_3}^{(2)} = {a_3}^{(2)}P_3 + {b_3}^{(2)}P_1.}
\end{array}
\end{equation}
We have the characteristic number on six points of a conic (CNC) as~\eref{eq:CNC} by substituting the representation coefficients $a_i^{(j)}$ and $b_i^{(j)}$ into~\eref{eq:CN}, and the CNC value equals $+1$. We apply this property of CN to pick arc segments likely lying on an ellipse.
\begin{equation}\label{eq:CNC}
CNC({\cal P},{\cal Q}) = \prod\limits_{i = 1}^3 {\prod\limits_{j = 1}^2 {\frac{b_i^{(j)}}{a_i^{(j)}}} } = 1.
\end{equation}

We provide a simple proof to $CNC({\cal P},{\cal Q})=+1$ based on Pascal's hexagon theorem~\cite{stefanovic2010very} as below.

\textbf{Proof}
Let $\{{Q_i}^{(j)}|i=1,2,3;j=1,2\}$ be six points on a conic. As shown in~\fref{fig:ProveCNC}, we can obtain three more intersections as
\begin{equation}\label{eq:defineR3}
\left\{ \begin{gathered}
  {R_1} =  < Q_2^{(2)}Q_3^{(1)},P_1P_2 > , \hfill \\
  {R_2} =  < Q_1^{(1)}Q_3^{(2)},P_2P_3 > , \hfill \\
  {R_3} =  < Q_1^{(2)}Q_2^{(1)},P_1P_3 > . \hfill \\
\end{gathered}  \right.
\end{equation}

Then $R_1$, $R_2$, and $R_3$ can be represented by $\{{Q_i}^{(j)}|i=1,2,3;j=1,2\}$ and $\{P_1,P_2,P_3\}$ through simple calculations as
\begin{equation}\label{eq:eq2}
\left\{ \begin{gathered}
  {R_1} =  - |{Q_2}^{(2)},{Q_3}^{(1)},{P_2}|{P_1} + |{Q_2}^{(2)},{Q_3}^{(1)},{P_1}|{P_2},\\
  {R_2} =  - |{Q_1}^{(1)},{Q_3}^{(2)},{P_3}|{P_2} + |{Q_1}^{(1)},{Q_3}^{(2)},{P_2}|{P_3},\\
  {R_3} =  - |{Q_1}^{(2)},{Q_2}^{(1)},{P_3}|{P_1} + |{Q_1}^{(2)},{Q_2}^{(1)},{P_1}|{P_3},\\
\end{gathered}  \right.
\end{equation}
where we use homogeneous coordination to represent a planar
point as $A=[A(x),A(y),A(z)]^T$, and $|A,B,C|$ denotes the determinant of the $3\times3$ matrix given by the homogeneous coordinates of the three points $A$, $B$ and $C$ as
\begin{equation}
|A,B,C| = \left| {\begin{array}{*{20}{c}}
  {A(x)}&{B(x)}&{C(x)} \\
  {A(y)}&{B(y)}&{C(y)} \\
  {A(z)}&{B(z)}&{C(z)}
\end{array}} \right|.
\end{equation}

As the homogeneous coordinate of a point in a projective plane is independent on the initial points constructing the plane, we specify $P_1=(1,0,0)^T, P_2=(0,1,0)^T$, and $P_3=(0,0,1)^T$. Consequently, the points $R_1$,$R_2$ and $R_3$ can be represented by $P_i$, $a_i^{(j)}$ and $b_i^{(j)} (i=1,2,3;j=1,2)$ by substituting~\eref{eq:p} into~\eref{eq:eq2} as
\begin{equation}\label{eq:q}
\left\{ {\begin{array}{*{20}{l}}
  {{R_1} =  ( b_2^{(2)}b_3^{(1)},- a_2^{(2)}a_3^{(1)},0)^T,} \\
  {{R_2} =  (0, - b_1^{(1)}b_3^{(2)},a_1^{(1)}a_3^{(2)})^T,} \\
  {{R_3} = ( a_1^{(2)}a_2^{(1)},0,- b_1^{(2)}b_2^{(1)})^T.}
\end{array}} \right.
\end{equation}

According to Pascal's hexagon theorem~\cite{stefanovic2010very}, $R_1$,$R_2$, and $R_3$ are collinear, i.e.,
\begin{equation}\label{eq:sameline}
  |R_1,R_2,R_3|=0.
\end{equation}
The proof of~\eref{eq:sameline} is provided in Appendix A.

Finally, we can obtain~\eref{eq:CNC} by substituting~\eref{eq:q} into~\eref{eq:sameline}. The proof is completed. Again, as seen from~\eref{eq:eq2}, the ratios of several determinants of $3\times3$ matrices generate CNC.

\section{Fast ellipse detection}
In this section, we present our ellipse detector using the characteristic number to prune non-elliptical line segments and pick arc segments lying on an ellipse. These pruning and picking processes significantly reduce the number of arc candidates for final fitting, rendering fast detection. The complete detection procedure includes preprocessing, line pruning, arc selection, and parameter fitting and ellipse validation.

At the preprocessing step, edge points are detected and linked to generate arc segments, where short segments are removed as noise. We delete those arc segments likely to be lines detected by CNL values at the line pruning step. This step reduces the number of arc segments that are not parts of any ellipses. However, the possible arc combinations are still too many to efficiently fit elliptical parameters. At the following arc selection step, we firstly divide arc segments into four groups, and remove some impossible arc combinations across these groups according to their relatively positional relationships. Subsequently, we apply CNC to pick the arc combinations belonging to one ellipse. Only those picked arc segments are used to fit elliptic parameters in the last fitting and validation step. The pruning and picking with CNL and CNC significantly reduce the number of possible arc combinations that determine the computational load of the fitting. Also, these pruning and picking steps run fast so that the overall detector is quite efficient.

\subsection{Preprocessing}
Given an image, we firstly smooth the image to partially suppress noise, and apply the Canny edge detector~\cite{canny1986computational} with default thresholds in Matlab to extract consecutive edge points. The edge detector outputs both the position ${x_i}$ and ${y_i}$, and gradient ${\tau _i}$ on each edge point as ${e_i} = ({x_i},{y_i},{\tau _i})$, where $i=1,2,...,N$, ${\tau _i}$ = $dy/dx$, $N$ is the number of edge points.

It is possible to apply the CN constraint on any three or six edge points to determine whether they lie in a line or ellipse. However, we have to calculate $C_{N}^{3}$ and $C_{N}^{6}$ point combinations for $N$ edge points, resulting in high computational complexity. Moreover, most of these combinations come from noise, or different ellipses and lines, spending a large amount time on invalid CN calculations. Instead, we apply the constraints on \emph{arc} segments.

\begin{figure*}
    \centering
    \subfigure[] { \includegraphics[width=0.2\textwidth]{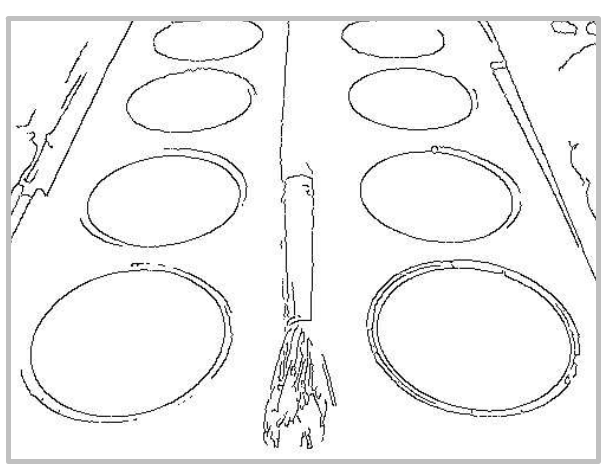} \label{fig:selectarc1} }
    \subfigure[] { \includegraphics[width=0.2\textwidth]{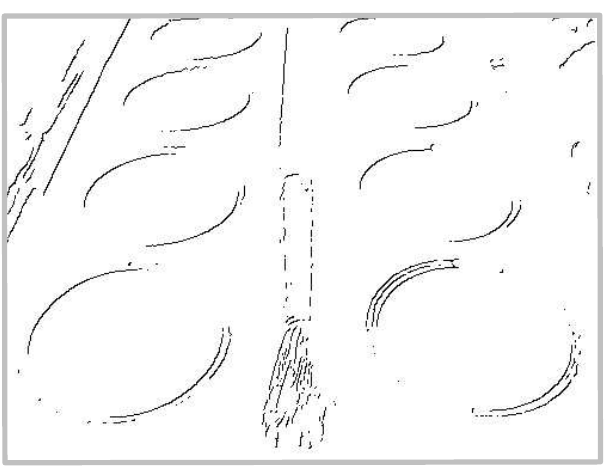} \label{fig:selectarc2} }
    \subfigure[] { \includegraphics[width=0.2\textwidth]{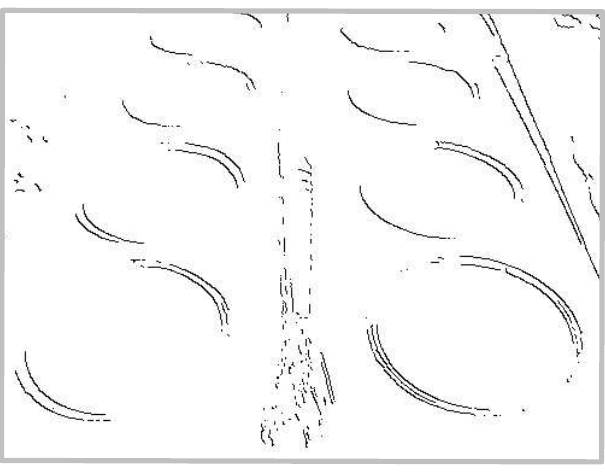} \label{fig:selectarc3} }
    \subfigure[] { \includegraphics[width=0.2\textwidth]{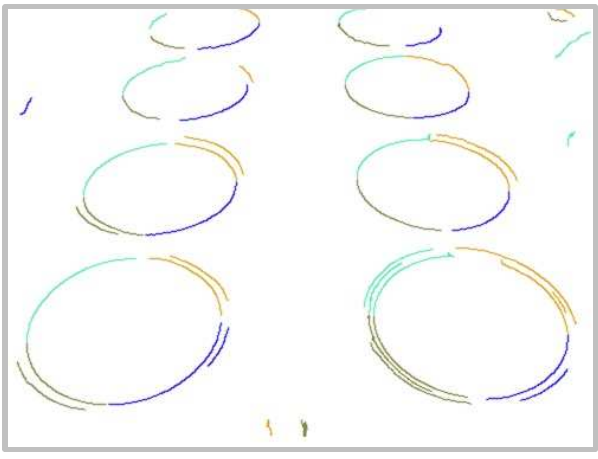} \label{fig:selectarc4} }
    \caption{Arcs detection and grouping. (a) shows edge points from Canny detector; (b) shows the arcs in sets $Arc_{\uppercase\expandafter{\romannumeral2}}$ and $Arc_{\uppercase\expandafter{\romannumeral4}}$; (c) shows the arcs in sets $Arc_{\uppercase\expandafter{\romannumeral1}}$ and $Arc_{\uppercase\expandafter{\romannumeral3}}$; (d) shows the result after removing noise and lines, and the arcs in the same set are labeled in the same color. There are four colors represent arcs in four sets.} \label{fig:selectarc}
\end{figure*}

In order to efficiently shear invalid arc combinations for later processing, we group arc segments into four sets corresponding to the arcs from one ellipse distributing in four quadrants as $Arc_
{\uppercase\expandafter{\romannumeral1}}$, $Arc_{\uppercase\expandafter{\romannumeral2}}$, $Arc_
{\uppercase\expandafter{\romannumeral3}}$, and $Arc_{\uppercase\expandafter{\romannumeral4}}$, named \emph{quadrant sets}. In the preprocessing step, we separate edge points $e_i (i=1,2,...,N)$ into two groups $Arc_{\uppercase\expandafter{\romannumeral2}} \cup Arc_{\uppercase\expandafter{\romannumeral4}}$ and $Arc_{\uppercase\expandafter{\romannumeral1}} \cup Arc_{\uppercase\expandafter{\romannumeral3}}$ by the signs of edge gradients ${\tau_i}$ as the first stage of this arc grouping:

\begin{equation}\label{eq:fourParts1}
\left\{ \begin{array}{l}
\tau_i > 0,{e_i} \in Arc_{\uppercase\expandafter{\romannumeral2}} \cup Arc_{\uppercase\expandafter{\romannumeral4}},\\
\tau_i < 0,{e_i} \in  Arc_{\uppercase\expandafter{\romannumeral1}} \cup Arc_{\uppercase\expandafter{\romannumeral3}}.
\end{array} \right.
\end{equation}
We link each edge point with the other edge points in its eight neighborhood from the same group with a breadth-first strategy until no edge point exists in the neighborhoods. Consequently, we separately generate a series of arc segments in two groups shown in~\fref{fig:selectarc}. Figure~\ref{fig:selectarc2} demonstrates the arcs by linking edge points from the Canny detector (shown in~\fref{fig:selectarc1}) of the group $Arc_{\uppercase\expandafter{\romannumeral2}} \cup Arc_{\uppercase\expandafter{\romannumeral4}}$, and~\fref{fig:selectarc3} shows the arc segments of $Arc_{\uppercase\expandafter{\romannumeral1}} \cup Arc_{\uppercase\expandafter{\romannumeral3}}$.

\begin{figure}
\centering
\includegraphics[height=1.2in]{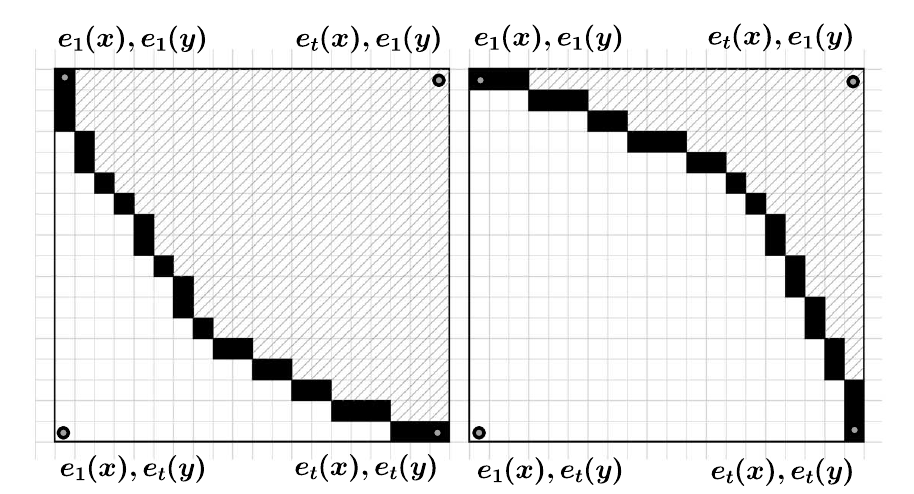}
\includegraphics[height=1.2in]{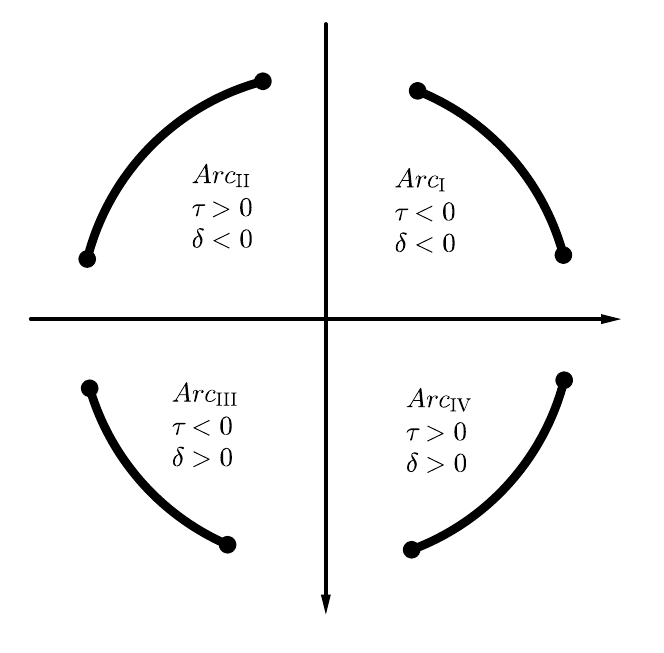}
\caption{Grouping arcs into four sets. The first two figures are two arcs in $Arc_{\uppercase\expandafter{\romannumeral1}} \cup Arc_{\uppercase\expandafter{\romannumeral3}}$. The difference between the numbers of pixels above and below (slashed and solid white blocks) can split the sets into two sets further. The last figure shows the splitting process.}
\label{fig:aboveBelow}
\end{figure}

As the second stage of arc grouping, we divide each group into two sets, eventually producing four sets. As shown in~\fref{fig:aboveBelow}, the vertices $(e_1(x), e_1(y))$, $(e_1(x), e_t(y))$, $(e_t(x), e_1(y))$, and $(e_t(x), e_t(y))$ constitute the bounding box of an arc with the length of $t$, where the starting and ending edge points of the arc are $e_1$ and $e_t$, respectively. Denoting the difference between the numbers of pixels above and below (slashed and solid white blocks in~\fref{fig:aboveBelow}) as $\delta$, we split each group into two by the sign of $\delta$. Taking an arc segment in the $Arc_{\uppercase\expandafter{\romannumeral1}} \cup Arc_{\uppercase\expandafter{\romannumeral3}}$ group for example, the arc falls in the $Arc_{\uppercase\expandafter{\romannumeral3}}$ set if the pixel number above the arc is greater than the below ($\delta > 0$) otherwise ($\delta < 0$) falls in the $Arc_{\uppercase\expandafter{\romannumeral1}}$ set. The last plot in~\fref{fig:aboveBelow} illustrates this splitting processing. We preform this stage of grouping after line pruning, and hence the cases of $\delta=0$ are quite rare. Finally, every arc segment in each of these four sets comes from one unique ellipse so that only arc combinations from different sets are necessary for the arc picking process. Therefore, this preprocessing is also designated for an efficient ellipse detector.

\subsection{Line pruning} \label{sec:line_pruning}

\begin{figure}
    \centering
    \includegraphics[width=0.45\textwidth]{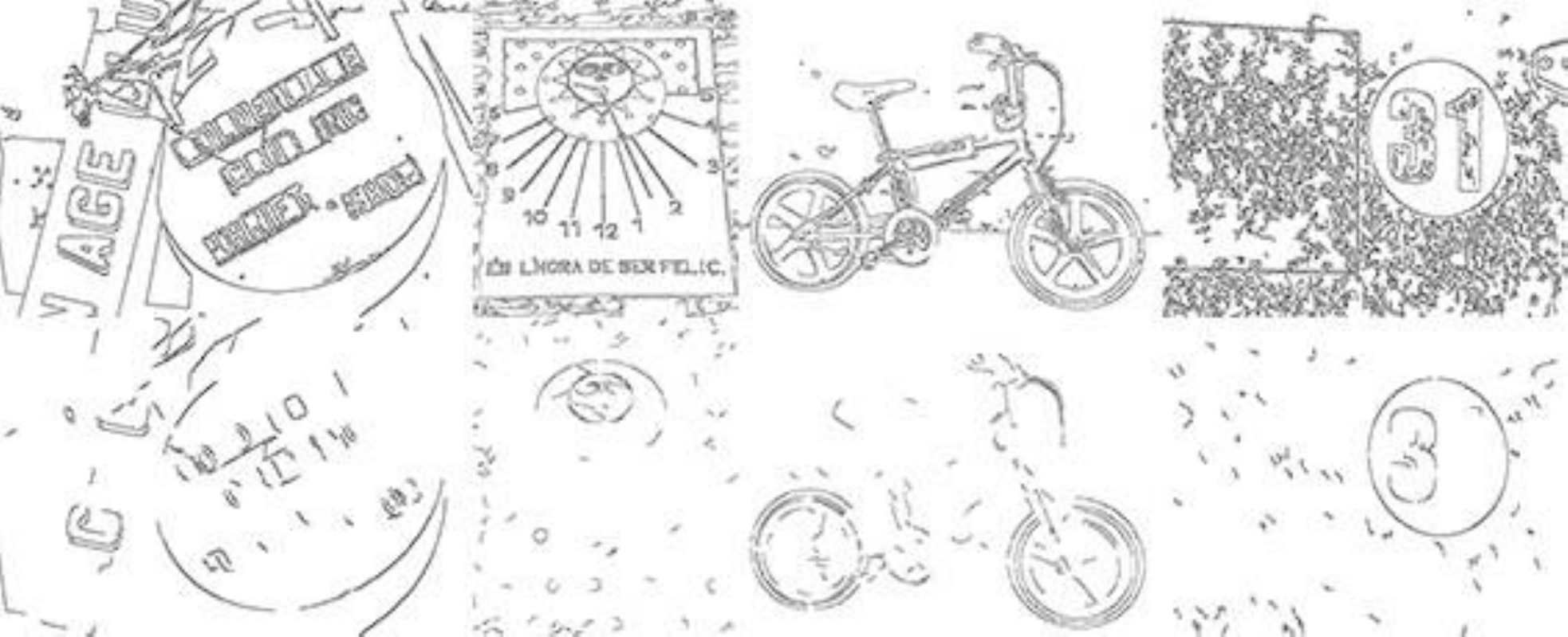}
    \caption{The segments after removing noise and lines. The top row gives the arc segments with edge linking, and the bottom provides those after noise and line pruning.} \label{fig:linePruning}
\end{figure}

As a typical treatment to noise in literature, we simply set a threshold $Th_{length}$ ($Th_{length}=16$ in all experiments of this study) to remove short segments. Following the noise attenuation, we employ CNL to prune line segments unlikely to come from an ellipse. The determinant of three edge points, the starting, midpoint, and ending of an arc segment ($e_1, e_i$ and $e_t$) is taken as the fast calculation of the CNL value for the segment given in~\eref{eq:sameline}. The geometric interpretation of the determinant is the area of $\triangle e_1e_ie_t$. We use the ratio of the area to the length of the arc ($CNL/t$) to robustly estimate whether these three points are collinear. The ratio close to zero indicates a line segment to be removed. Figures~\ref{fig:selectarc4} and~\ref{fig:linePruning} presents the segments after removing noise and lines. The top row of~\fref{fig:linePruning} gives the arc segments with edge linking, and the bottom provides those after noise and line pruning. Many segments from noise and lines disappear, significantly reducing the number of arc candidates.

\subsection{Arc selection}\label{sec:arc_selection}

\begin{figure}
    \centering
    \includegraphics[width=0.40\textwidth]{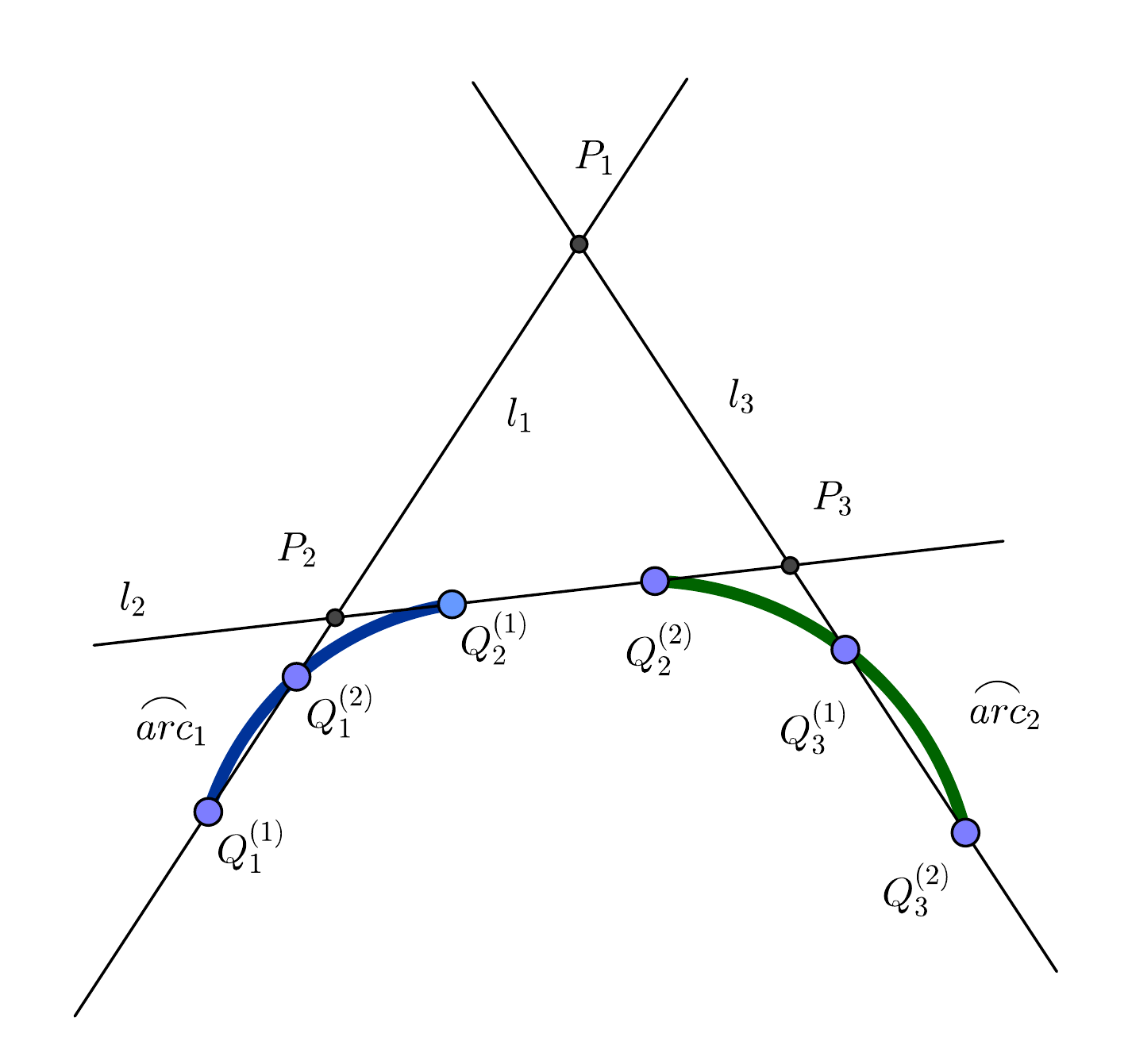}
    \caption{Construction of CNC based on two arcs. $Q_1^{(2)}$ is the mid points of $\stackrel \frown{arc_1}$, while $Q_1^{(1)}$ and $Q_2^{(1)}$ are two endpoints of $\stackrel \frown{arc_1}$. Similarly, $Q_3^{(1)}$, $Q_2^{(2)}$ and $Q_3^{(2)}$ are three points on $\stackrel \frown{arc_2}$.}
    \label{fig:howToUse}
\end{figure}

\begin{figure}
    \centering
    \includegraphics[width=0.40\textwidth]{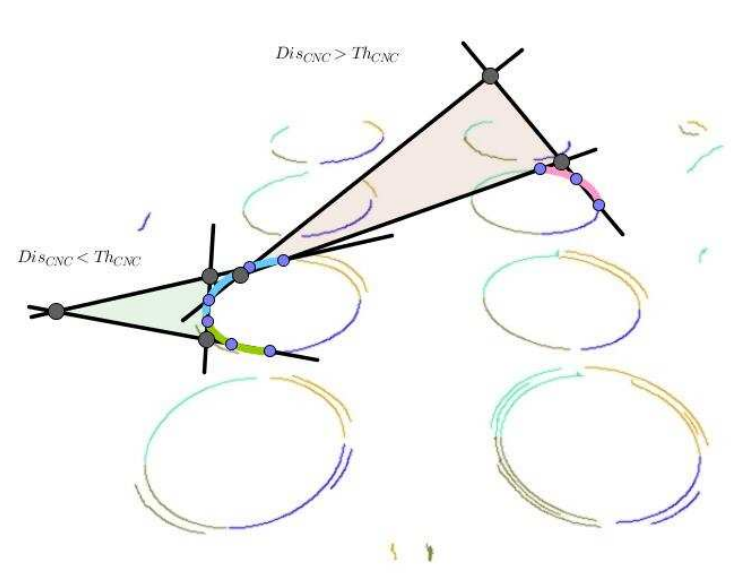}
    \caption{Threshold ($Th_{CNC}$) is used to determine false negative for arc selection. The arc pair (labeled in green and blue) belonging the same ellipse obtains the value of CNC, $Dis_{CNC}$, close to +1, and the triangle is labeled in light green. The red-brown triangle indicates the arc pair (labeled in blue and pink) having the CNC value greater than $Th_{CNC}$. Points on arcs are labeled in purple, and vertices of triangles are labeled in grey.}
    \label{fig:CNC_Performance}
\end{figure}

In this step, we pick candidate arcs, that are likely to assemble an ellipse, across the four sets where noise and line segments have been removed. Specifically, two arcs, each with three edge points, are taken from two different sets (one arc per set) in order to construct CNC.
As shown in~\fref{fig:howToUse}, there are two arcs $ \stackrel \frown{arc_1}$ and $\stackrel \frown{arc_2}$, where $Q_1^{(2)}$, $Q_1^{(1)}$ and $Q_2^{(1)}$ are the mid and two endpoints of $\stackrel \frown{arc_1}$, respectively. Similarly, $Q_3^{(1)}$, $Q_2^{(2)}$ and $Q_3^{(2)}$ are the points in $\stackrel \frown{arc_2}$. We intersect $Q_1^{(1)}Q_1^{(2)}$ and $Q_3^{(1)}Q_3^{(2)}$ at $P_1$, $Q_2^{(1)}Q_2^{(2)}$ and $Q_1^{(1)}Q_1^{(2)}$ at $P_2$, and $Q_2^{(1)}Q_2^{(2)}$ and $Q_3^{(1)}Q_3^{(2)}$ at $P_3$. Given these points in $\mathcal{P}$ and $\mathcal{Q}$, we are able to have the representation coefficients $a_i^{(j)}$ and $b_i^{(j)}$ ($i=1,2,3$ and $j=1,2$), and calculate the CNC value for this arc combination by substituting these coefficients into~\eref{eq:CNC}. Equation~\ref{eq:CNC} also tells that the CNC value equals $+1$ if these two arc segments with six points come from one identical ellipse. Therefore, picking two arc candidates from one ellipse turns out to simply comparing the CNC value of the arc combination with the value $+1$. Similar to line pruning, we can also use a threshold ($Th_{CNC}$) that determines false negative for this process. The arc pair belonging the same ellipse obtains the value of CNC close to $+1$, and the absolute difference of CNC value and $+1$, represented as $Dis_{CNC}$, is close to 0, as shown in~\fref{fig:CNC_Performance} where the triangle constructed by the six points from the pair is labeled in light green. The red-brown triangle indicates the arc pair having the CNC value greater than $Th_{CNC}$, where the two arc segments of this pair lie on different ellipses.

We only consider the combinations with arc segments from two adjacent quadrant sets, e.g., $\{Arc_{\uppercase\expandafter{\romannumeral1}}, Arc_{\uppercase\expandafter{\romannumeral2}}\}$ and $\{Arc_{\uppercase\expandafter{\romannumeral2}}, Arc_{\uppercase\expandafter{\romannumeral3}}\}$, but leave out those two-arc combinations from diagonal sets $\{Arc_{\uppercase\expandafter{\romannumeral1}}, Arc_{\uppercase\expandafter{\romannumeral3}}\}$ and $\{Arc_{\uppercase\expandafter{\romannumeral2}}, Arc_{\uppercase\expandafter{\romannumeral4}}\}$. This choice further lowers down possible combinations, and more importantly improves the robustness to calculate CNC values. As shown in~\fref{fig:howToUse}, the lines $Q_1^{(1)}Q_1^{(2)}$ and $Q_3^{(1)}Q_3^{(2)}$ are likely to be parallel for arcs from diagonal quadrant sets so that the intersection $P_1$ does not exit, or lies quite far from the edge points. The calculation of CNC value is so instable that the determination of ellipse is no more effective. We exclude the combinations with arcs from two diagonal quadrant sets, avoiding this instability.

\begin{figure}[t]
\centering
    \subfigure{\includegraphics[width=0.24\textwidth]{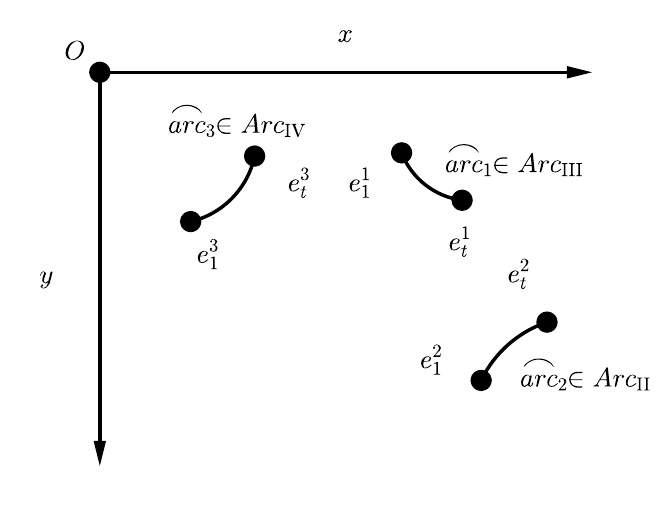}}
    \subfigure{\includegraphics[width=0.24\textwidth]{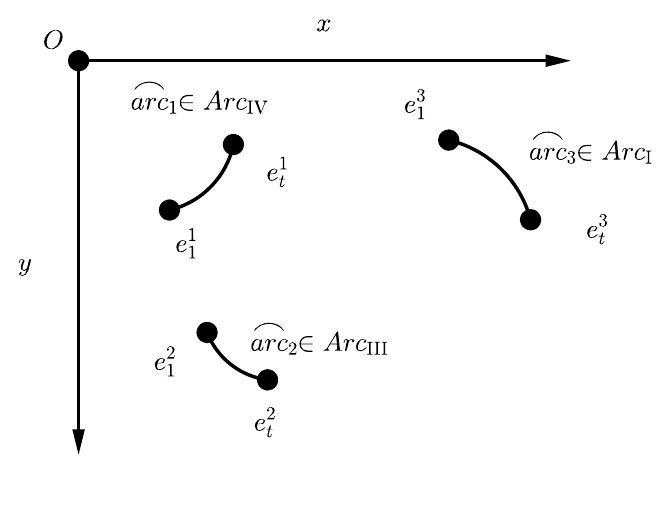}}
    \caption{Two examples of invalid three-arc combinations that form one ellipse.}
    \label{fig:errorPosition}
\end{figure}

The combinations with more arc segments are also able to yield more accurate parameter estimation in the later step. Unfortunately, noise and/or occlusions are likely to bring the absence of arc segments from one quadrant set. Hence, we constitute the combinations with~\emph{three} arcs from three different quadrant sets upon the arc pairs picked by CNC for later parameter fitting. There are four kinds of valid arc combinations from three quadrant sets for an ellipse, as shown in the first row of~\tref{tab:position}. We pick out these combinations in each kind by using both coordinate and CNC constraints together, where the coordinate constraint takes relative locations of arcs to remove invalidate combinations. The second and third rows of~\tref{tab:position} list the picking rules for valid combinations using CNC and coordinate constraints, respectively. The symbols $e_1^1$ and $e_t^1$ denote the starting and end points of $\stackrel \frown{arc_1}$.  Figure~\ref{fig:errorPosition} illustrates two examples of invalid three-arc combinations that form one ellipse, removed by the constraints in~\tref{tab:position}.

\begin{table*}[htbp]
  \centering
  \caption{The picking rules for valid combinations using CNC and coordinate constraints}
    \begin{tabular}{m{0.1\textwidth}<{\centering}
    m{0.2\textwidth}<{\centering}m{0.2\textwidth}<{\centering}
    m{0.2\textwidth}<{\centering}m{0.2\textwidth}<{\centering}}
    \toprule
    Valid arc combinations
    &{\includegraphics[width=0.2\textwidth]{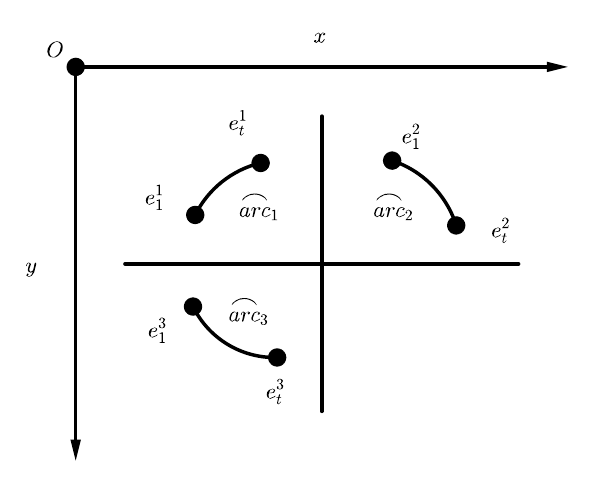}}&{\includegraphics[width=0.2\textwidth]{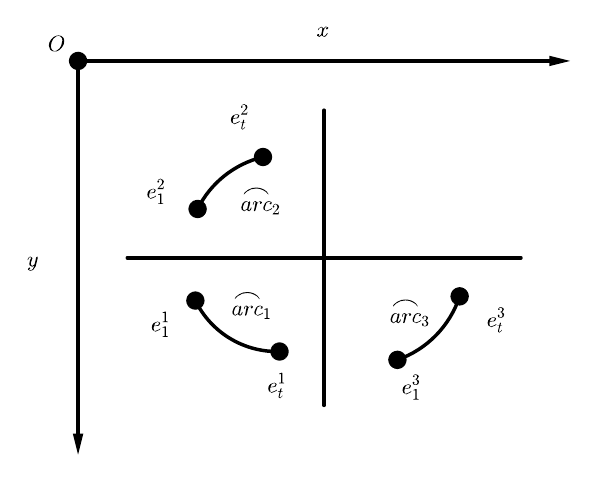}}&{\includegraphics[width=0.2\textwidth]{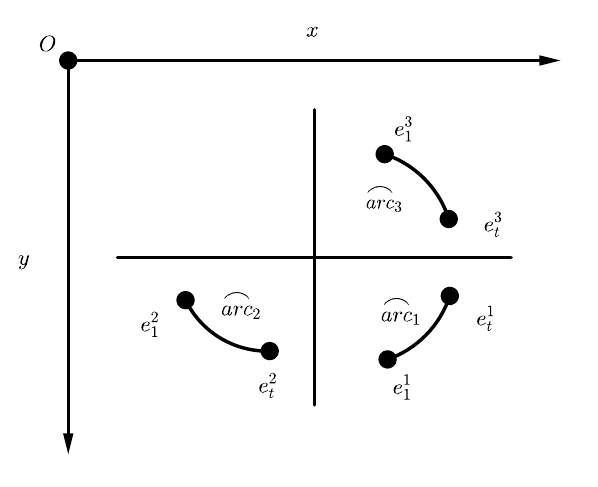}}&{\includegraphics[width=0.2\textwidth]{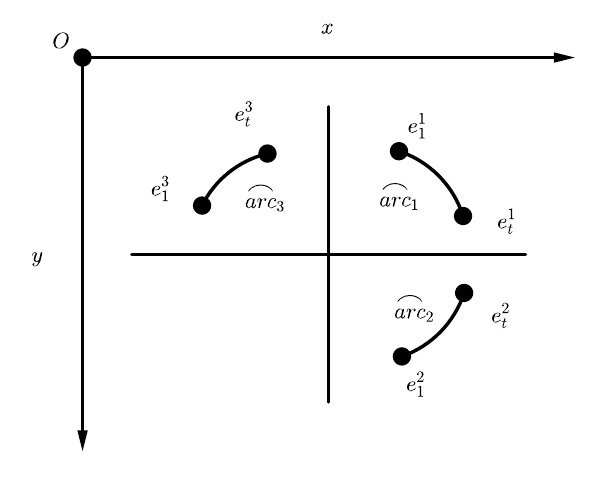}}\\
    \midrule
    \\
    Sets constraints & $\left\{ \begin{array}{l}{
          \wideparen{arc}}_{1} \in {Arc}_{\uppercase\expandafter{\romannumeral2}}\\
          {\wideparen{arc}}_{2} \in {Arc}_{\uppercase\expandafter{\romannumeral1}}\\
          {\wideparen{arc}}_{3} \in {Arc}_{\uppercase\expandafter{\romannumeral3}}
        \end{array} \right.$
        & $\left\{ \begin{array}{l}{
          \wideparen{arc}}_{1} \in {Arc}_{\uppercase\expandafter{\romannumeral3}}\\
          {\wideparen{arc}}_{2} \in {Arc}_{\uppercase\expandafter{\romannumeral2}}\\
          {\wideparen{arc}}_{3} \in {Arc}_{\uppercase\expandafter{\romannumeral4}}
        \end{array} \right.$
        & $\left\{ \begin{array}{l}{
          \wideparen{arc}}_{1} \in {Arc}_{\uppercase\expandafter{\romannumeral4}}\\
          {\wideparen{arc}}_{2} \in {Arc}_{\uppercase\expandafter{\romannumeral3}}\\
          {\wideparen{arc}}_{3} \in {Arc}_{\uppercase\expandafter{\romannumeral1}}
        \end{array} \right.$
        & $\left\{ \begin{array}{l}{
          \wideparen{arc}}_{1} \in {Arc}_{\uppercase\expandafter{\romannumeral1}}\\
          {\wideparen{arc}}_{2} \in {Arc}_{\uppercase\expandafter{\romannumeral4}}\\
          {\wideparen{arc}}_{3} \in {Arc}_{\uppercase\expandafter{\romannumeral2}}
        \end{array} \right.$\\
    \\
    \midrule
    \\
    Coordinate constraints &$\left\{ \begin{array}{l}
    {e_t^1(x)} < {e_1^2(x)}\\
          {e_1^1(y)} < {e_1^3(y)}\end{array} \right.$
          & $\left\{ \begin{array}{l}{e_t^1(x)} < {e_1^3(x)}\\
          {e_1^1(y)} > {e_1^2(y)}\end{array} \right.$
          & $\left\{ \begin{array}{l}{e_1^1(x)} > {e_t^2(x)}\\
          {e_t^1(y)} > {e_t^3(y)}\end{array} \right.$
          & $\left\{ \begin{array}{l}{e_1^1(x)} > {e_t^3(x)}\\
          {e_t^1(y)} < {e_t^2(y)}\end{array} \right.$\\
          \\
    \bottomrule
    \end{tabular}%
  \label{tab:position}%
\end{table*}%

Specifically, the picking process for a three-arc combination begins with an arc segment in the middle quadrant set, and then proceed to those in the other two sets. We find an arc pair first, and then the third arc to form the combination by alternatively applying the coordinate and CNC constraints. We take the set combination $\{Arc_{\uppercase\expandafter{\romannumeral1}}, Arc_{\uppercase\expandafter{\romannumeral2}}, Arc_{\uppercase\expandafter{\romannumeral3}}\}$ in~\tref{tab:position} as an example to illustrate the picking process. We starts with one arc in $Arc_{\uppercase\expandafter{\romannumeral2}}$, and test the pair of the arc of $Arc_{\uppercase\expandafter{\romannumeral2}}$ and every arc segments in $Arc_{\uppercase\expandafter{\romannumeral1}}$ with the coordinate constraint. If one pair meet the coordinate constraint, the CNC constraint is applied to this pair further. Subsequently, we search the set $Arc_{\uppercase\expandafter{\romannumeral3}}$ to find the third arc segments forming a pair with the arc in $Arc_{\uppercase\expandafter{\romannumeral2}}$ that follow the coordinate constraint. Herein, the CNC constraint runs only once in order to balance the speed and accuracy. We repeatedly run the picking process for every arc segments in $Arc_{\uppercase\expandafter{\romannumeral2}}$, and find all valid three-arc combinations for $\{Arc_{\uppercase\expandafter{\romannumeral1}}, Arc_{\uppercase\expandafter{\romannumeral2}}, Arc_{\uppercase\expandafter{\romannumeral3}}\}$. Algorithm 1 details the picking process for this set combination. The similar process applies to the other three-set combinations in the first row of~\tref{tab:position}, and finally we have all valid three-arc combinations ready for parameter fitting given in the next section.

\begin{algorithm}[H]
\caption{Picking algorithm for a three-arc combination}
    \begin{algorithmic}
    \renewcommand{\algorithmicrequire}{\textbf{Input:}}
    \renewcommand{\algorithmicensure}{\textbf{Output:}}
    \REQUIRE ${Arc}_{\uppercase\expandafter{\romannumeral1}}$
        ${Arc}_{\uppercase\expandafter{\romannumeral2}}$
        ${Arc}_{\uppercase\expandafter{\romannumeral3}}$
    \ENSURE  valid arc combinations set
    \FOR {\textbf{each} ${\wideparen{arc}}_{i}\in{{Arc}_{\uppercase\expandafter{\romannumeral2}}}$}
        \FOR {\textbf{each} ${\wideparen{arc}}_{j} \in {{Arc}_{\uppercase\expandafter{\romannumeral1}}}$}
            \IF {${\wideparen{arc}}_{i},{\wideparen{arc}}_{j}$ do not meet coordinate constraints }
                \STATE Continue.
            \ENDIF
            \IF {$Dis_{CNC}({\wideparen{arc}}_{i},{\wideparen{arc}}_{j}) > {Th}_{CNC}$ }
                \STATE Continue.
            \ENDIF
            \FOR {\textbf{each} ${\wideparen{arc}}_{k} \in {{Arc}_{\uppercase\expandafter{\romannumeral3}}}$}
                \IF {${\wideparen{arc}}_{i},{\wideparen{arc}}_{k}$ do not meet coordinate constraints}
                    \STATE Continue.
                \ENDIF
                \STATE
                Add $\{{\wideparen{arc}}_{i} {\wideparen{arc}}_{j} {\wideparen{arc}}_{k}\} $ to valid arc combinations set
            \ENDFOR
        \ENDFOR
    \ENDFOR
    \RETURN valid arc combinations set
    \end{algorithmic}
\end{algorithm}

\subsection{Parameter fitting}
There are five parameters to determine an ellipse, including its center (two), orientation (one), and major and minor semi-axes (two). We follow the procedure in~\cite{fornaciari2014fast} to estimate all these parameters but the center whose calculation we provide below.

\begin{figure}[t]
\centering
    \subfigure
    {\includegraphics[width=0.35\textwidth]{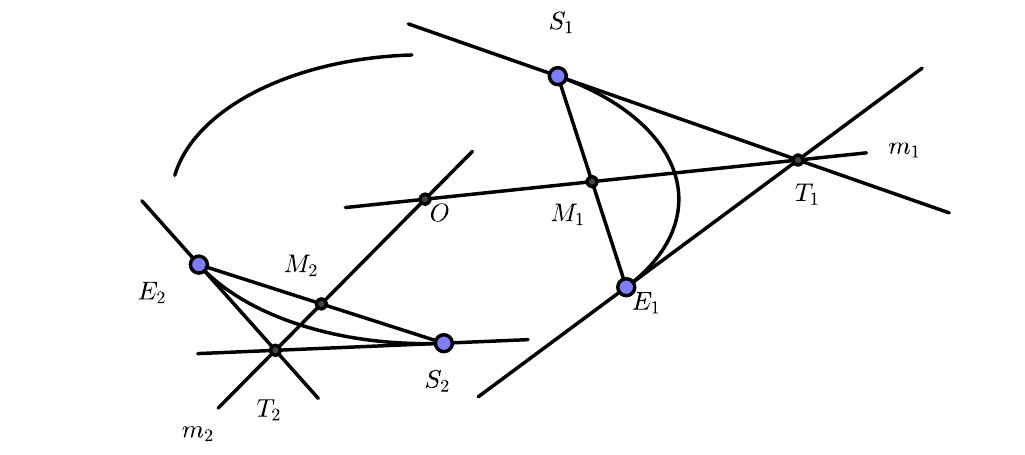}}
    \subfigure
    {\includegraphics[width=0.35\textwidth]{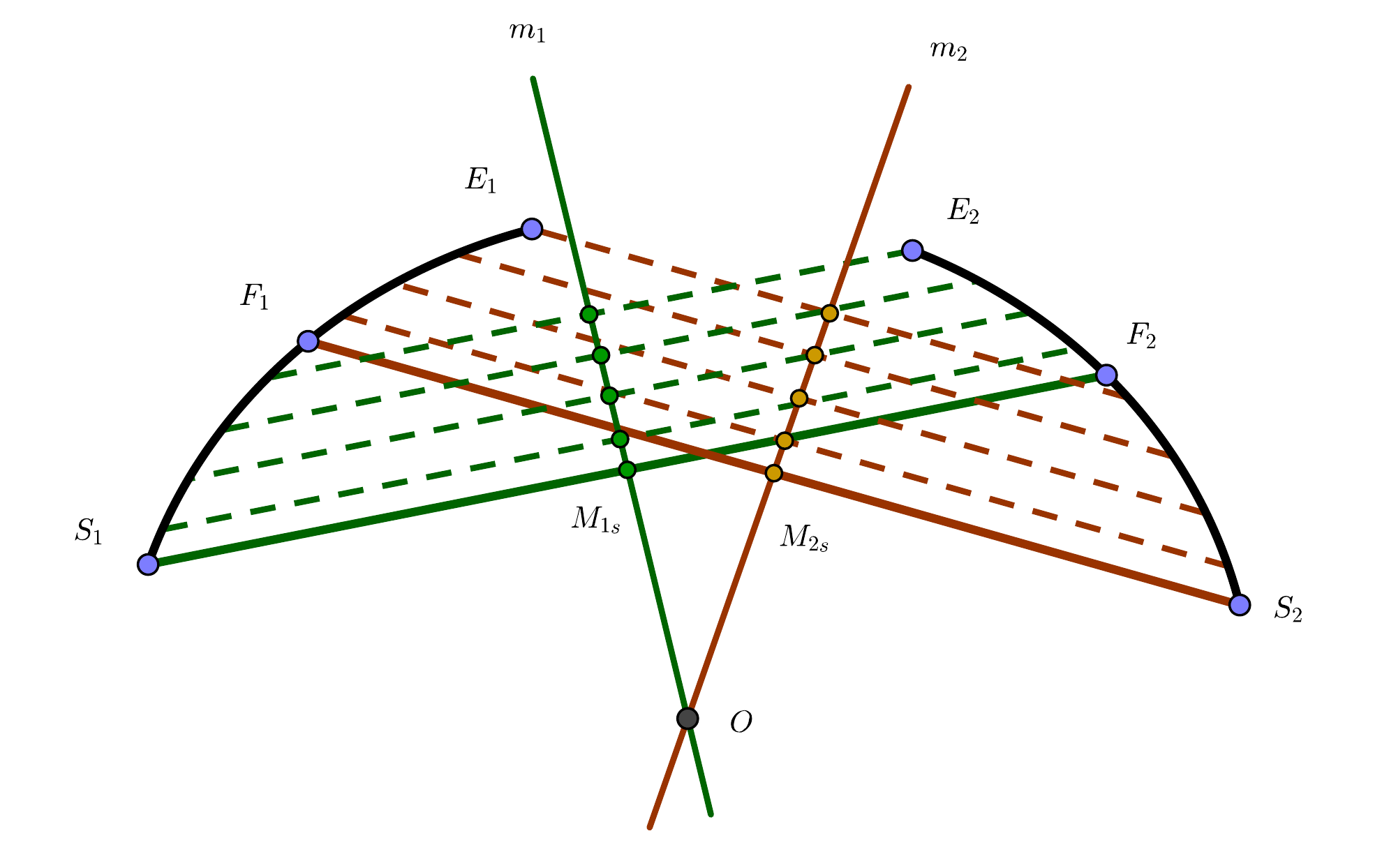}}
    \caption{Computing the center of the ellipse. The top figure shows the ellipse center calculated by two arcs. The bottom figure shows the ellipse center calculated by a series of parallel chords.}
     \label{fig:ellipseFit1}
\end{figure}

We estimate the center as the intersection of auxiliary lines generated from any two-point pairs on the three arcs picked by the previous procedure. As illustrated in the top figure of ~\fref{fig:ellipseFit1}, $S_1$ and $E_1$ are two points on an arc segment, $T_{1}$ is the intersection of the two tangent lines from $S_1$ and $E_1$, and $M_{1}$ is the middle point of the line segment $S_1E_1$. It is proved that the line $M_{1}T_{1}$ passes through the ellipse center $O$. For practical images, it is unnecessary that all such auxiliary line segments like $M_{1}T_{1}$ given by any point pairs on the elliptical arcs, but we are able to locate the center as the point closest to those lines by the least square fitting. In this study, we find that the accuracy of this fitting (estimation) also depends on the tangents of the points along the ellipse.

We use $n_d$ ($n_d=16$ for our experiments) parallel chords instead of tangent lines in order to minimize the effects from tangent deviations, as shown in the bottom figure of ~\fref{fig:ellipseFit1}. The points $F_1$ and $F_2$ are the mid points of the arcs $\wideparen{S_1E_1}$ and $\wideparen{S_2E_2}$, respectively. We generate $n_d$ chords parallel to $S_1F_2$, and so we do to the chord $S_2F_1$. The points $M_{1_s}$ and $M_{2_s}$ ($s=1,2,...,{n_d}$) are the mid-points of the two series of parallel chords to $S_1F_2$ and $S_2F_1$, respectively. The points of $M_{1_s}$ are collinear at $m_1$, $M_{2_s}$ lying on the line $m_2$, and the intersection of $m_1$ and $m_2$ determines the ellipse center $O$. We estimate $m_1$ and $m_2$ using a fast variant of the robust Theil-Sen estimator~\cite{matouvsek1991randomized} with two arcs in adjacent quadrant sets. Consequently, we obtain four lines through the ellipse center generated from a three-arc combination, yielding at most six pairwise intersections. The algebraic average of the coordinates of these six intersections is taken as the output ellipse center $(x_0,y_0)$.

\subsection{Ellipses validation}
Candidate ellipses are available after the parameter fitting step. There may exist false positives or duplicated ones in these candidates so that a validate step is necessary.

We provide two indices to validate the candidates. The first one measures how many edge points fitting the corresponding ellipse. The more edge points fit, the more likely the ellipse exists. We substitute every edge points into the correspondence ellipse equation to calculate how it fits the equation. We count the number of the fitting edge points, and calculate the ratio to the total points that generate the ellipse by fitting arc combinations.

The second index accounts for arc lengths of three-arc combinations giving ellipse parameters. Ideally, the length of an arc combination is larger than the sum of two semi-axes. We use the ratio between the total length of one three-arc combination and three times of the semi-axes summation as the index. Larger values indicate more stable detection of the ellipse.

The validation in~\cite{prasad2010clustering} is able to discover duplicated ellipses, which allows to assess the similarity of two ellipses by comparing the differences of ellipse parameters. A voting strategy is adopted to pick up the center of a given cluster, removing duplicate ellipses.

\section{Experimental results and analysis}
We perform a series of experiments on data sets with both synthetic and real images to evaluate the performance of our fast detector by comparing with other state-of-the-art methods~\footnote{The source code and resultant images of our detector can be found at https://github.com/dlut-dimt/ellipse-detector}.

\subsection{Evaluation metrics and data sets}

All the experiments in this paper are executed on a desktop with Intel(R) Core(TM) i7-6700 CPU whose clock frequency is $3.40$ GHz. The performance of ellipse detectors is evaluated in terms of running time and F-measure. The most time-consuming step in many existing methods lie in parameter fitting because too many arc or point candidates are fed to the step. In addition to executing time, we count candidate combinations of arcs (CC) for ellipse fitting, and use CC as a direct index to demonstrate the impacts of CNC and CNL.

F-measure is defined as:
\begin{equation}\label{fmeasure}
{\rm{F-measure}} = \frac{{2 \times {\rm{Precision}} \times {\rm{Recall}}}}{{{\rm{Precision + Recall}}}},
\end{equation}
where
\begin{equation}\label{pr}
{\rm{Precision}} = \frac{\Psi}{\Omega}, {\rm{Recall}} =\frac{\Psi}{\Gamma}.
\end{equation}
The symbol $\Omega$ denotes the number of detected ellipses, and $\Gamma$ indicates the number of ground-truth ellipses. $\Psi$ is the number of correctly detected ellipses. The overlapping ratio of a detected ellipse ${\cal{E}}_d$ to the ground truth ${\cal{E}}_g$ is defined as:
\begin{equation}
{\cal{M}}({{\cal{E}}_d},{{\cal{E}}_g}) = \frac{{area({{\cal{E}}_d}) \cap area({{\cal{E}}_g})}}{{area({{\cal{E}}_d}) \cup area({{\cal{E}}_g})}},
\end{equation}
where $area(\cal{E})$ is the number of pixels inside the ellipse ${\cal{E}}$. The detected ellipse ${\cal{E}}_d$ is considered as a correct detection if ${\cal{M}}({{\cal{E}}_d},{{\cal{E}}_g})>Th_{o}$. The threshold $Th_{o}$ is set to 0.8 throughout our experiments, as did in~\cite{prasad2012edge}.

Experimental data contain both real world and synthetic images. We use data sets with real world images to validate efficiency and accuracy, while synthetic ones are designated to demonstrate robustness to noise and ellipse variations.
Real world images are those from Dataset \#1 and Dataset Prasad, the same as~\cite{fornaciari2014fast}, for fair comparisons. Dataset \#1 is composed of 400 images having elliptic shapes, collected from MIRFlickr and LabelMe repositories~\cite{hartley2003multiple}. Those MIRFlickr images are of high quality, and most of them contain only one object (ellipse), while those from LabelM are noisy images of low resolution, containing multiple objects. Dataset Prasad contains 198 real images from~\cite{prasad2012edge}, where objects of oval shapes like human faces are regarded as ellipses. Besides evaluations on the original Dataset Prasad, we also construct Dataset \#2 by selecting 50 images with rigorous ellipses in order to show the accuracy of our method on detecting ellipses. These rigorously elliptical shapes are quite common in industrial inspection and diagnosis.

\subsection{Performance analysis}
The effectiveness of line pruning and arc selection steps are illustrated by comparisons on detection results with and without these steps. We also provide empirical studies on the hyper-parameters involved in these steps. The parameters for ellipse fitting and validation are taken the same values as those in~\cite{fornaciari2014fast}.

\subsubsection{Performance analysis of line pruning}
We remove short and straight line segments using CNL as discussed in Section~\ref{sec:line_pruning}. Our goal is to prune noise effects and lines in input images as much as possible but preserve arcs of ellipses. We perform line pruning on Dataset \#1 by tuning $Th_{CNL}$ from $0$ to $5.0$. Intuitively, the threshold $Th_{CNL}$ limits the height of the triangle formed by the three edge points for CNL calculation. Therefore, the larger $Th_{CNL}$ is, the more arcs are to be removed. The zero threshold, $Th_{CNL}=0$, indicates no line pruning step included.

Table~\ref{tab:addlabel} lists the values of arc numbers, running time, and F-measure with varying $Th_{CNL}$ values. There are averagely $181$ arc segments without line pruning ($Th_{CNL}=0$), and more segments are removed when increasing $Th_{CNL}$ values. The computing time is $28.86$ms, and F-measure is $0.4313$ without line pruning. When noise and line segments are removed by CNL, the computing time is becoming lower and F-measure increases. We obtain the best performance, $8.52$ms for computation and $0.4692$ for F-measure when $Th_{CNL}=3.0$. The computing time is still decreasing when more arc segments are pruned by the thresholds larger than $3.0$. The values of F-measure slightly decrease due to increased false negative. These results show that the line pruning step using CNL is effective to remove noise and lines, and also alleviate the computational load for later steps. We set $Th_{CNL}=3.0$ to balance the accuracy and computing time in all comparisons with the others.

\begin{table}[htbp]
  \centering
  \caption{The performance of line pruning step.}
    \begin{tabular}{rrrr}
    \toprule
    \parbox{1.2cm}{$Th_{CNL}$} & arc num & avg. time(ms) & F-measure \\
    \midrule
    0     & 181   & 28.86  & 0.4313  \\
    1     & 170   & 26.52  & 0.4338  \\
    2     & 108   & 13.75  & 0.4576  \\
    \textbf{3} & \textbf{65 } & \textbf{8.52 } & \textbf{0.4692 } \\
    4     & 40    & 6.79  & 0.4499  \\
    5     & 26    & 6.17  & 0.4226  \\
    \bottomrule
    \end{tabular}
  \label{tab:addlabel}
\end{table}

\subsubsection{Performance analysis of arc selection}
We use the geometric constraint of six points on arcs derived from CNC to pick up arc segments belonging to one ellipse. Theoretically, the CNC value of six points lying on an ellipse equals $+1$, but various imaging conditions (e.g., thermal noise and lens distortions) in practical applications may cause the value deviating from $+1$. As discussed in Section~\ref{sec:arc_selection}, we relax this hard constraint to a range in the vicinity of $+1$ determined by $Th_{CNC}$. Herein, we perform experimental analysis on the relationship between point coordinates and CNC values. This analysis does not only give arise to an appropriate threshold, but also validates the effectiveness of the arc selection based on CNC.

\begin{figure}
\centering
\includegraphics[width=0.45\textwidth]{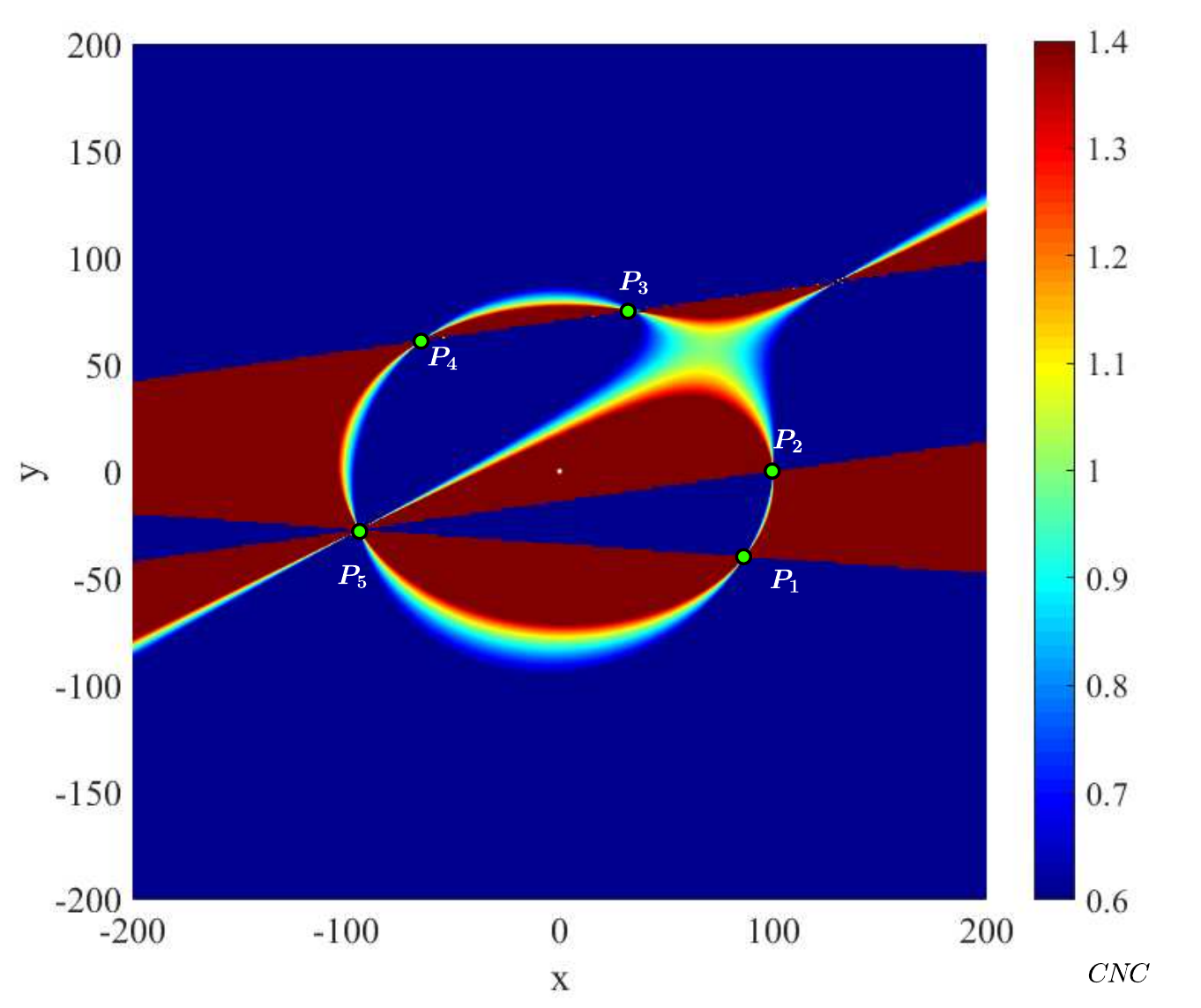}
\caption{Relationship between point coordinates and CNC values. Different colors indicate various CNC values given by the color bar on the right most. Five distinct points $P_1,...,P_5$ on the ellipse are fixed, and the sixth point around the ellipse are varied to show the distribution of CNC values.}
\label{fig:BOEperformance}
\end{figure}

Supposing an ellipse centering at the coordinate origin, we fix five distinct points $P_1,...,P_5$ on the ellipse, and vary the sixth point around the ellipse to show the distribution of CNC values in~\fref{fig:BOEperformance}. Different colors indicate various CNC values given by the color bar on the right most of~\fref{fig:BOEperformance}. All the CNC values higher than $1.4$ are colored in red-brown, while all those lower than $0.6$ in blue. Figure~\ref{fig:BOEperformance} illustrates that most of CNC values with the sixth point close to the ellipse fall within the range from $0.6$ to $1.4$. There exist several regions where CNC values lie in the range while the sixth point locates far from the ellipse. These regions include the star-shaped area out of the ellipse between $P_2$ and $P_3$, and the bottom left and top right ends of the line stretching out of $P_5$.  This observation can be explained by the fact that the CNC value of $+1$ indicates the points lying on a conic curve not only an ellipse. However, these 'outliers' have few effects on our detector, because other types of conic curves, e.g., hyperbola and parabola, seldom appear in practical industrial images, and the majority can be also removed by the coordinate constraint even if a few of them appear.

\begin{figure*}
\centering
\includegraphics[width=0.9\textwidth]{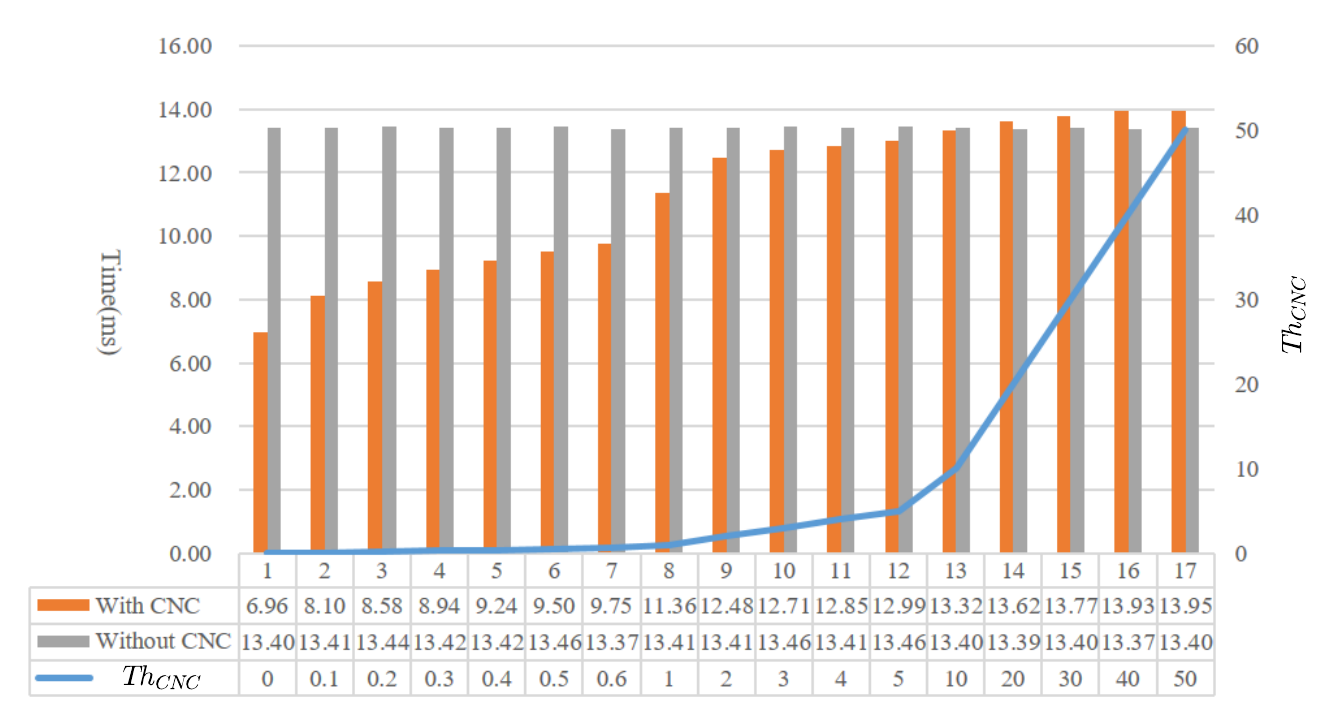}
\caption{The values of computing time with varying thresholds. We vary the absolute deviation from $+1$, from $0$ to $50$, giving seventeen $Th_{CNC}$ values in total, labeled in blue line and listed in the third row. The orange and grey bars show the computing time with and without the arc selection step.}
\label{fig:CNCTime}
\end{figure*}

We further apply our detector on dataset \#1 by varying $Th_{CNC}$, the absolute deviation from $+1$, from $0$ to $50$, giving seventeen values in total. Figure~\ref{fig:CNCTime} illustrates the values of computing time with varying thresholds. The blue line in~\fref{fig:CNCTime} demonstrates the seventeen values of $Th_{CNC}$ in an ascending order, also listed in the third row below the plot. The step values vary with the value ranges of $Th_{CNC}$, i.e., $0.1$ for the range from $0$ to $0.6$, $1.0$ for $1$ to $5$, and $10$ for values larger than $10$. The orange and grey bars show the computing time with and without the arc selection step, whose values are listed in the first and second rows below, respectively. The average computing time is $6.96$ ms for each image, one half of the detection time without CNC, $13.4$ms, when we use the hard constraint $Th_{CNC}=0$ picking only a small fraction of arc segments. More arc segments are included for parameter fitting, demanding more computing time, when increasing the threshold values. When $Th_{CNC}$ is larger than $20$, the detector spends more time than that without the selection step since the calculation of CNC takes more time than what the constraint can save. As a result, the orange bars are higher than the grey ones for these $Th_{CNC}$ values in~\fref{fig:CNCTime}.

\begin{figure*}
\centering
\includegraphics[width=0.9\textwidth]{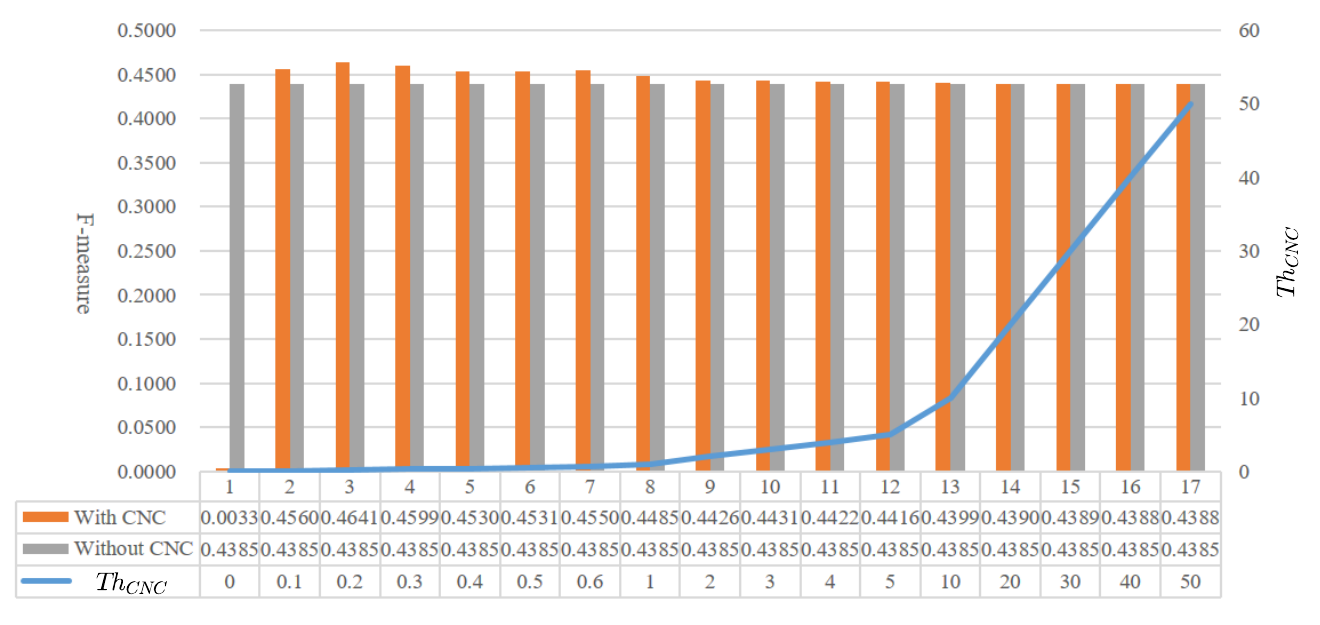}
\caption{The values of F-measure (F-m) with varying thresholds. We take the same deviation from $+1$ as did in \fref{fig:CNCTime}, while the orange and grey bars show the F-measure with and without the arc selection step.}
\label{fig:CNCFm}
\end{figure*}

Figure~\ref{fig:CNCFm} shows the values of F-measure (F-m) with varying threshold values. The F-measure for the detector without the CNC constraint is $0.4385$ labeled in the gray bar, while ours in orange bars. The hard constraint yields a very low F-m $0.0033$ since the choice excludes many arc segments slightly deviating from an ellipse, resulting in significant false negatives. The threshold $0.2$ outputs the best F-m $0.4641$, even higher than the detector without this constraint. This improvement shows that the CNC constraint is also able to exclude false positives in addition to decreasing the computing time from $13.40$ms to $8.58$ms. As expected, the values of F-m given by the thresholds larger than $20$ are quite close to that without the selection step. In these cases, the selection step takes no effect on lowering down false positives. Therefore, the arc selection step using CNC is quite crucial to both efficiency and accuracy. We choose the threshold as $Th_{CNC}=0.2$ to generate the best performance in the following experiments.

\subsection{Comparisons with the state of the art}

Firstly, we compare our detector with three recent arc-based methods, i.e., Zhang~\cite{zhang2005robust}, Libuda~\cite{libuda2007ellipse}, and Fornaciari~\cite{fornaciari2014fast}, on Dataset \#1. The set consists of images with different qualities and various numbers of target ellipses. The latest works of~\cite{liang2015robust} and~\cite{mulletiellipse} are not so relevant as these three because both are point-based and applicable to sceneries where only one ellipse appear. The execution program of~\cite{fornaciari2014fast} is provided by the authors, and we take the results of~\cite{zhang2005robust} and~\cite{libuda2007ellipse} reported in~\cite{fornaciari2014fast}. Table~\ref{tab:compareOthers} lists the comparisons on average running time and F-measure. The method of Zhang~\cite{zhang2005robust} performs the worst: the fitting on a large number of pixel combinations takes much time over $4$s for an image; its F-measure is the lowest as $0.3421$. Libuda~\cite{libuda2007ellipse} and Fornaciari~\cite{fornaciari2014fast} output the similar efficiency with the average execution time for each image $14.38$ ms and $12.79$ ms, respectively. The values of their F-measure are also quite close. Our detector achieves the best in terms of both efficiency($8.54$ms) and accuracy (F-measure about $0.47$). Specifically, we buy 7\% higher accuracy with 33\% less execution time than the fastest method~\cite{fornaciari2014fast}.

\begin{table}[htbp]
  \centering
  \caption{F-measure and runtime compared with the state of the art on Dataset \#1}
	\begin{tabular}{rrr}
    \toprule
    Method & F-measure & avg.time(ms) \\
    \midrule
    Libuda\cite{libuda2007ellipse} & 0.4258 & 14.38  \\
    Prasad\cite{prasad2012edge} & 0.4512 & 823.38  \\
    Zhang\cite{zhang2005robust} & 0.3421 & 4243.86  \\
    Fornaciari\cite{fornaciari2014fast} & 0.4385  & 12.79  \\
    Our   & \textbf{0.4692 } & \textbf{8.54 } \\
    \bottomrule
    \end{tabular}%
  \label{tab:compareOthers}%
\end{table}

\begin{figure*}
\centering
\includegraphics[width=0.83\textwidth]{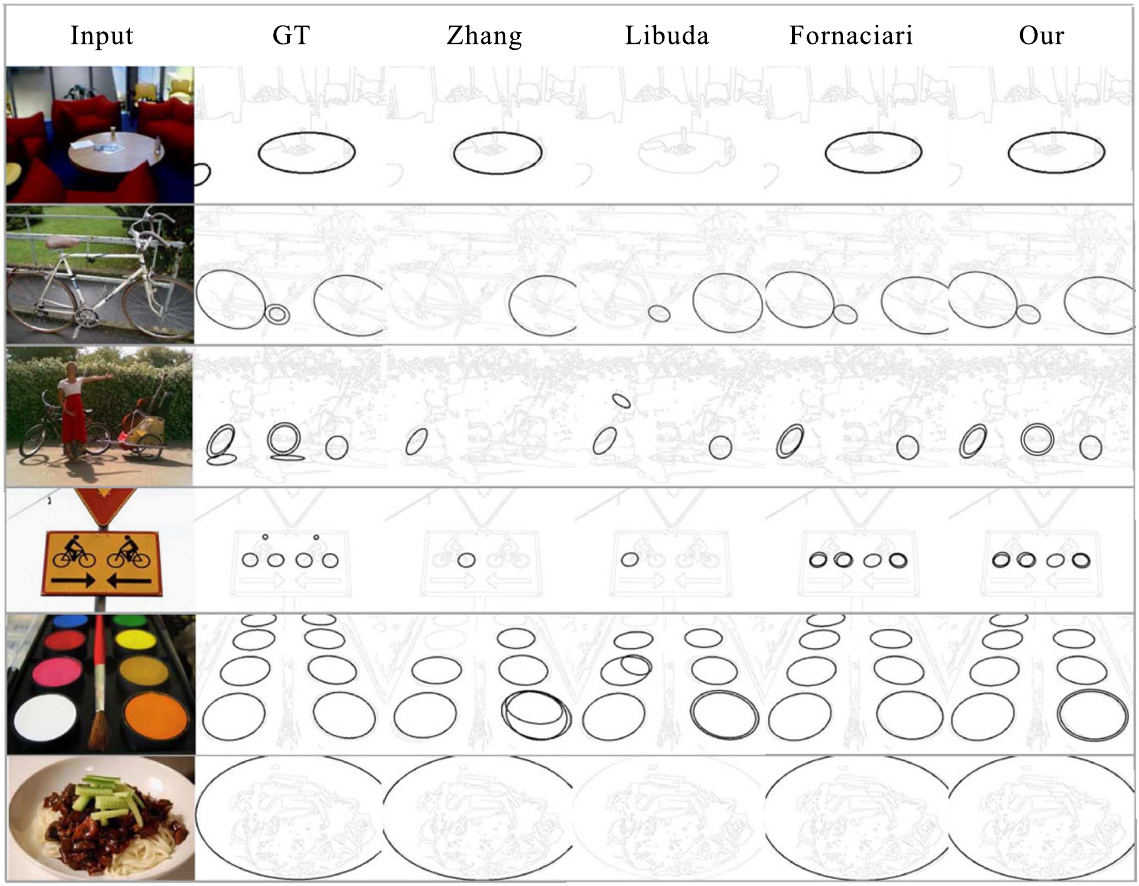}
\caption{Several examples of the compared methods on Dataset \#1. The first column is the input images, and the second one lists the ground truth (GT).}
\label{fig:compareOD2}
\end{figure*}

Several examples of the above methods on Dataset \#1 are shown in~\fref{fig:compareOD2} where the first column is the input images, and the second one lists the ground truth (GT). False negatives occur in all the results of Libuda~\cite{libuda2007ellipse} except for the fifth row. The method of Fornaciari~\cite{fornaciari2014fast} works quite close to ours that correctly detects almost all the ellipses in these images. It is worth noting that only our method successfully picks out the middle wheel partially occluded by the lady in the third image, while the other three fail. Compared with the ground truth, our detector outputs one extra ellipse out of the bottom right tray. Actually, one can find a dim elliptical trail along the tray, but the ground truth neglects it.

Secondly, we compare the execution time on each processing step with the-state-of-the-art on Dataset \#1, as shown in \tref{tab:stepTime}. The method of Zhang~\cite{zhang2005robust} spends about $4243$ ms on estimating the parameters due to huge numbers of possible point combinations. The method of Prasad~\cite{prasad2012edge} reduces the time on parameter estimation via grouping arcs with curvature and convexity, but the grouping spends additional $278$ms. The method of Libuda~\cite{libuda2007ellipse} uses an iterative strategy, and the time spent on each of the first three steps equals about $4.5$ms. Fornaciari's detector~\cite{fornaciari2014fast}, the fastest one among the existing, spends the most time $4.9$ms on grouping arcs with their relative locations against the other steps. In contrast, our method only uses about half of the execution time of~\cite{fornaciari2014fast} on the grouping step, and the total execution time is less than two third of~\cite{fornaciari2014fast}. Thus, our detector reduces the time for the grouping step, the bottleneck for efficient detectors.

\begin{table*}[htbp]
  \centering
  \caption{Execution times (ms) for each step compared with the-state-of-the-art on Dataset \#1.}
    \begin{tabular}{rrrrrrr}
    \toprule
          & Libuda \cite{libuda2007ellipse} & Prasad \cite{prasad2012edge}& Zhang \cite{zhang2005robust} & Fornaciari \cite{fornaciari2014fast} & Our \\
    \midrule
    Edage detection   & 4.49  & 3.54  & 3.97  & 3.45  & 3.43  \\
    Pre-processing   & 4.15  & 78.03  & 3.55  & 1.94  & 1.90  \\
    Grouping   & 4.89  & 278.01  & 0.25  & 4.90  & \textbf{2.53 } \\
    Estimation   & 0.84  & 3.40  & 4236.06  & 2.30  & \textbf{0.63 } \\
    Validation and Clustering  & 0.00  & 460.39  & 0.03  & 0.21  & \textbf{0.06 } \\
    Total   & 14.38  & 823.38  & 4243.86  & 12.79  & \textbf{8.54 } \\
    \bottomrule
    \end{tabular}
  \label{tab:stepTime}
\end{table*}

\begin{figure}[t]
\centering
    \includegraphics[width=0.45\textwidth]{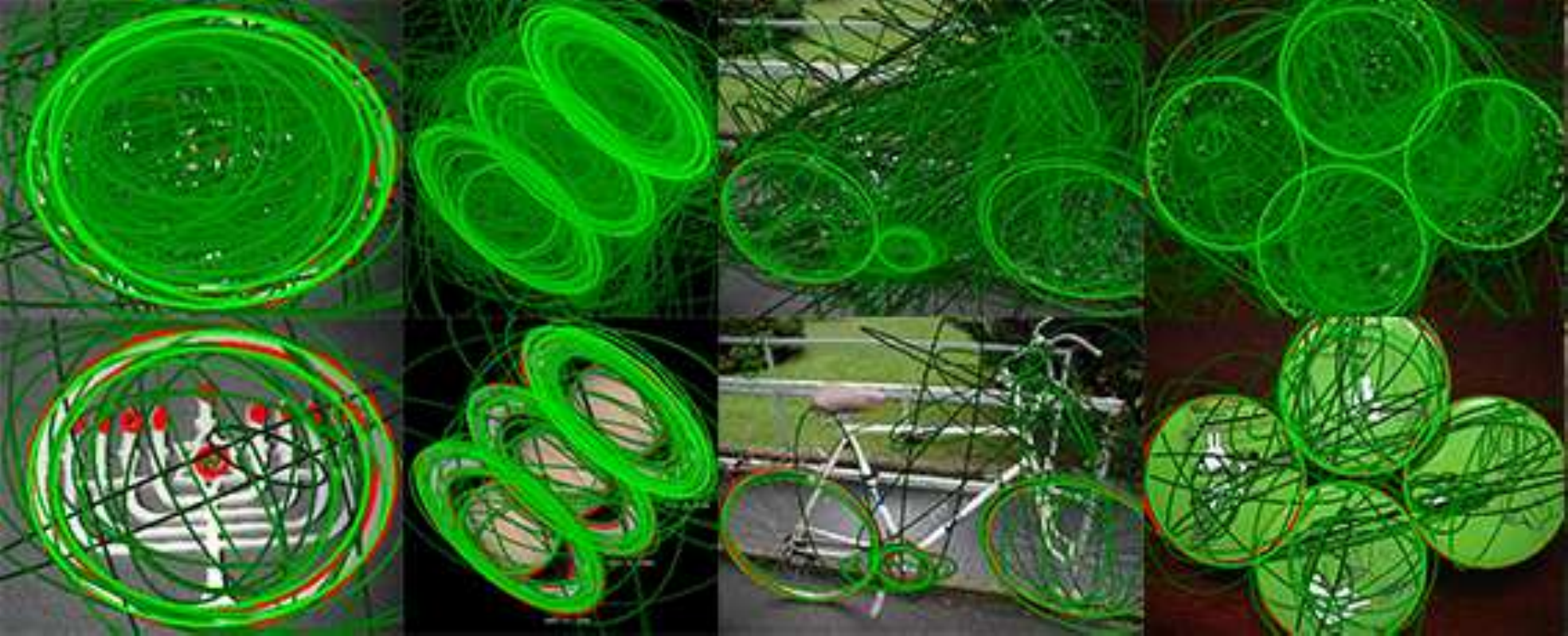}
    \caption{Ellipses needed to be fitted with or without line pruning and arc selection. Green curves in the first and second rows present possible ellipses to be fitted without and with our pruning/selection steps, respectively. }
    \label{fig:withCNC}
\end{figure}

Further, we peer into the proposed detector to analyze how line pruning and arc selection accelerate detection. Table~\ref{tab:pruningAndArcSelection} illustrates the effects of CNL and CNC by listing the averaged values of arc numbers, arc combination numbers (CC) for parameter fitting, execution time and F-measure on Dataset \#1. Without any processing using CNL nor CNC, possible averaged arcs and arc combinations (CC) are $181$ and $2619$, respectively. After line pruning, the arc number reduces to $65$, about one third of of the original, and the CC value significantly decreases to 196, $6\%$ of the original. Only $40$ arc combinations remain for parameter fitting, shearing $98.5\%$ combinations from the original. Naturally, the execution time drops down $93\%$ from $129.33$ms to $8.52$ms. The value of F-measure also increases from $0.3163$ to $0.4692$ since many false positives are removed by these two steps using CNL and CNC. Green curves in the first and second rows of~\fref{fig:withCNC} present possible ellipses to be fitted without and with our pruning/selection steps, respectively. It is evident that the green curves in the second row are much less than those in the first row, intuitively showing the effectiveness of our CNL/CNC based processing to save time and improve accuracy.

\begin{table}[htbp]
  \centering
  \caption{The effect of CNL and CNC}
    \begin{tabular}{m{0.1\textwidth}
    m{0.08\textwidth}<{\raggedleft}m{0.08\textwidth}<{\raggedleft}
    m{0.08\textwidth}<{\raggedleft}}
    \toprule
          & After arc detection & After line pruning & After arc selection \\
    \midrule
    arc Num & 181   & 65    & \textbf{65 } \\
    CC    & 2619  & 196   & \textbf{40 } \\
    avg.time (ms)  & 129.33  & 12.73  & \textbf{8.52 } \\
    F-measure   & 0.3163  & 0.4090  & \textbf{0.4692 } \\
    \bottomrule
    \end{tabular}
  \label{tab:pruningAndArcSelection}
\end{table}

Finally, we compare our detector with~\cite{fornaciari2014fast}, giving the best performance among the existing, in terms of CC, F-measure, and execution time on three data sets including Datast \#1, Dataset \#2, and Dataset Prasad. As shown in \tref{tab:CNCPerformance}, our method outperforms~\cite{fornaciari2014fast} on all the three data sets. The last column shows the improvement percentage both on efficiency and accuracy. The accuracy of the proposed detector increases about 6.6\% in average over~\cite{fornaciari2014fast}, and more significantly our detector averagely runs about 31.0\% faster than~\cite{fornaciari2014fast} on all the three data sets. These results show that our detector is applicable to real time scenarios in practice.

\begin{table*}[htbp]
  \centering
  \caption{Testing results of the proposed methods compared with \cite{fornaciari2014fast} on three datasets.}
    \begin{tabular}{rrrrrrrrrrr}
    \toprule
    \multirow{2}[3]{*}{DataBase} & \multicolumn{2}{c}{avg. time(ms)} & \multicolumn{2}{c}{F-m} & \multicolumn{2}{c}{CC} & \multicolumn{2}{c}{improvement percentage} \\
          & \cite{fornaciari2014fast} & Ours   & \cite{fornaciari2014fast} & Ours      & \cite{fornaciari2014fast}& Ours   & speed & accuracy \\
    \midrule
    Dataset Prasad & 4.34  & \textbf{3.35 } & 0.2874  & \textbf{0.3059 }    & 82    & 19    & 22.70\%  & 6.44\%  \\
    Dataset \#1 & 13.58  & \textbf{8.55 } & 0.4385  & \textbf{0.4692 }    & 267   & 40    & 37.05\%  & 7.00\%  \\
    Dataset \#2 & 5.26  & \textbf{4.09 } & 0.5893  & \textbf{0.6271 }   & 117   & 34    & 22.27\%  & 6.41\%  \\
    \bottomrule
    \end{tabular}%
  \label{tab:CNCPerformance}
\end{table*}%

\subsection{Robustness to ellipse variations and noise}
In order to investigate the robustness to ellipse variations and noise, we use two synthetic data sets with different orientations and ratios of two semi-axes of ellipses, and also apply salt-and-pepper noise to real world images. The noise break arcs into several small fragments, which may affect the accuracy and efficiency.

The first synthetic data set consists of $9100$ ellipses with various semi-axes ratios and orientations. One semi-axis is fixed as $100$, and the other one varies so that the ratios range from $0.01$ to $1$ at the step of $0.01$. Orientations vary from $1^\circ$ to $90^\circ$ at the step of $1^\circ$. The second data set contains $10000$ images with a fixed center and orientation, showing changes on ratios and lengths of semi-axes. One semi-axis varies from $1$ to $100$ at the step of $1$, and the other one changes accordingly so that axes ratios range from $0.01$ to $1$ at the step of $0.01$. Both data sets come from~\cite{fornaciari2014fast} that generates synthetic $400\times400$ images, each containing one single ellipse without noise.

\begin{figure*}
    \centering
    \subfigure[\cite{fornaciari2014fast}] { \includegraphics[width=0.23\textwidth]{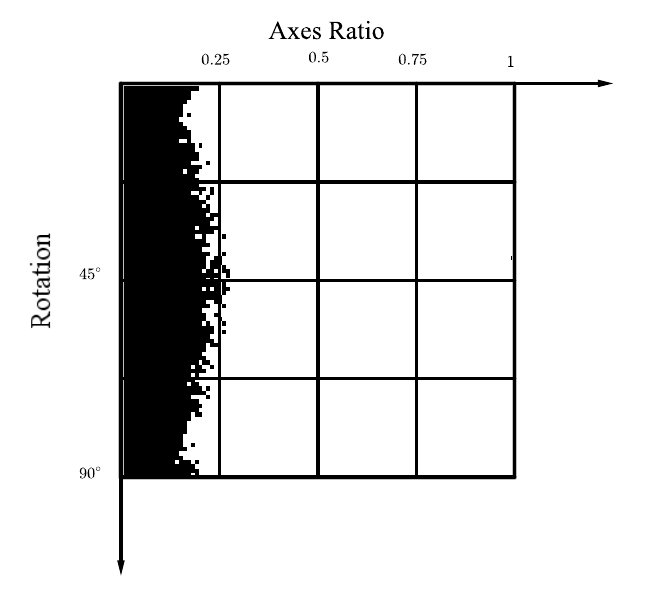} \label{fig:BRchange1} }
    \subfigure[Ours] { \includegraphics[width=0.23\textwidth]{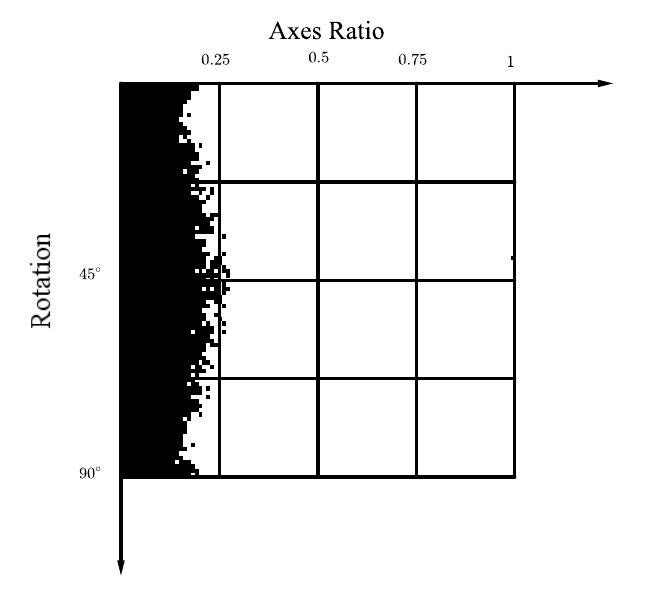} \label{fig:BRchange2} }
    \subfigure[\cite{fornaciari2014fast}] { \includegraphics[width=0.23\textwidth]{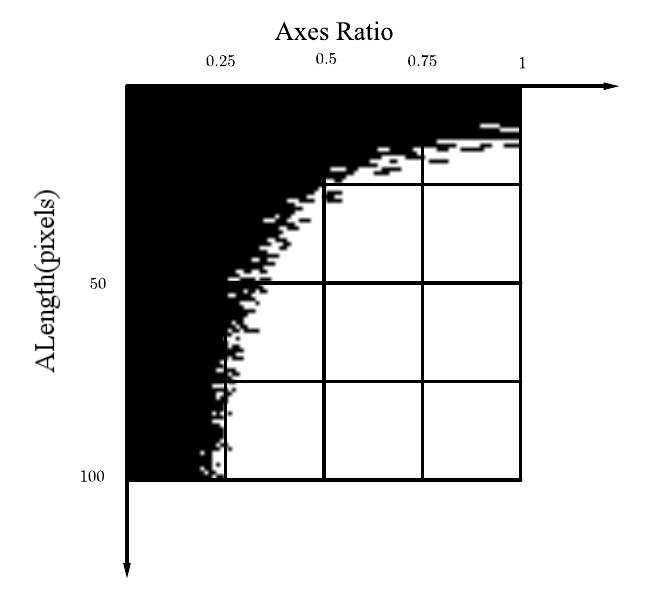} \label{fig:ABchange1} }
    \subfigure[Ours] { \includegraphics[width=0.23\textwidth]{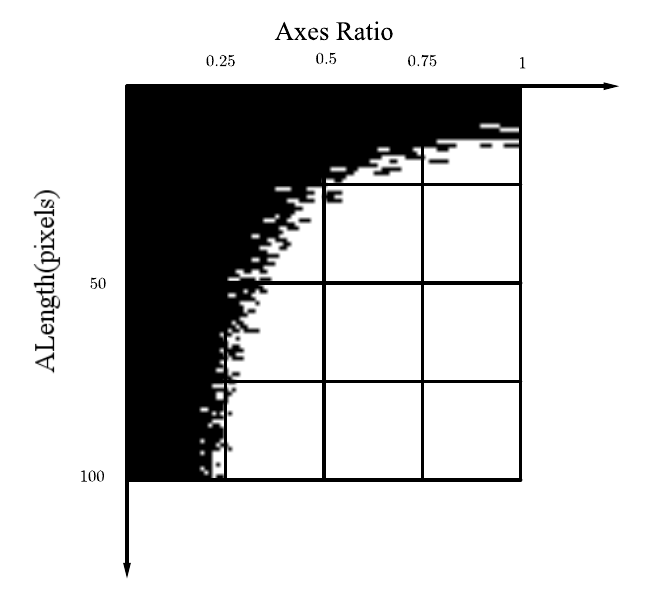} \label{fig:ABchange2} }
    \caption{Robustness to ellipse  variations compared with \cite{fornaciari2014fast}. Figure (a)-(b) are results on the first dataset. The horizontal axis gives ratios of two semi-axes, and the vertical one shows orientations. Figure (c)-(d) are results on the second dataset. The vertical axis indicates lengths of major axes, and the black points indicate failures of detection.}
    \label{fig:ABRchange}
\end{figure*}

Figure~\ref{fig:ABRchange} illustrates comparisons with~\cite{fornaciari2014fast}, where black points indicate failures of detection. The results on the first data set are given in \fref{fig:BRchange1} and \fref{fig:BRchange2}. The horizontal axis gives ratios of two semi-axes, and the vertical one shows orientations. Both methods are robust to orientation changes as long as axes ratios are larger than 0.25, but they fail in the cases of small axes ratios when ellipses degenerate into straight lines. The results on the second data set are shown in \fref{fig:ABchange1} and \fref{fig:ABchange2}, where the vertical axis indicates lengths of major axes. The robustness of both methods is quite similar, working well on ellipses whose major axis is longer than $10$ and axes ratio is larger than $0.25$. Small arcs are likely to be pruned as noise when the ellipse only has a few pixels. Fortunately, one may tackle the problem of small ellipses by upscaling the image as did in~\cite{fornaciari2014fast}. Therefore, the proposed method is quite robust, and only fails in some extreme cases, e.g., small and extremely oblate ellipses, which are quite rare in reality.

\begin{figure*}
    \centering
    \subfigure[F-measure] { \includegraphics[width=0.38\textwidth]{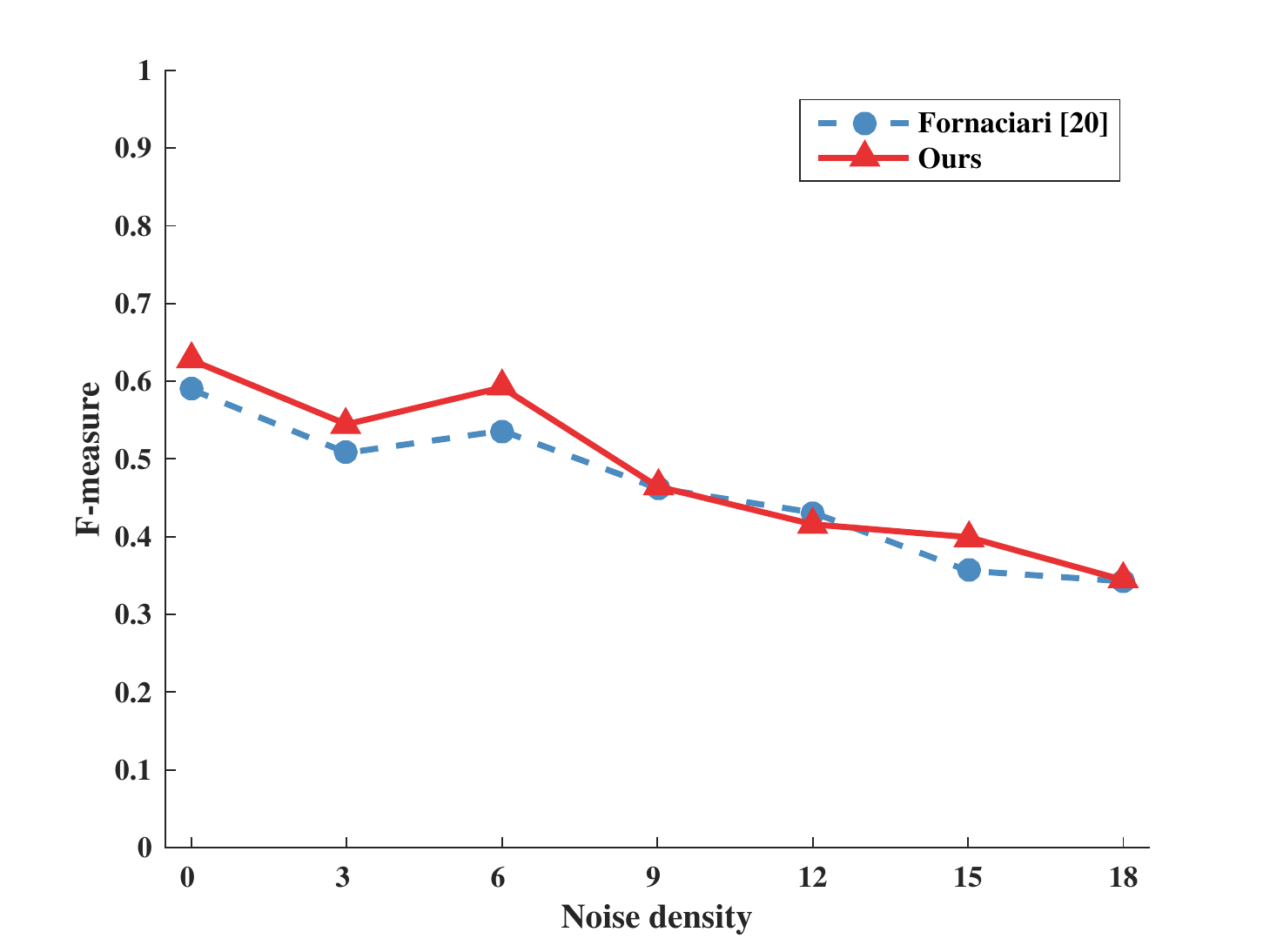} \label{fig:fmeasure} }
    \subfigure[Execution time] { \includegraphics[width=0.38\textwidth]{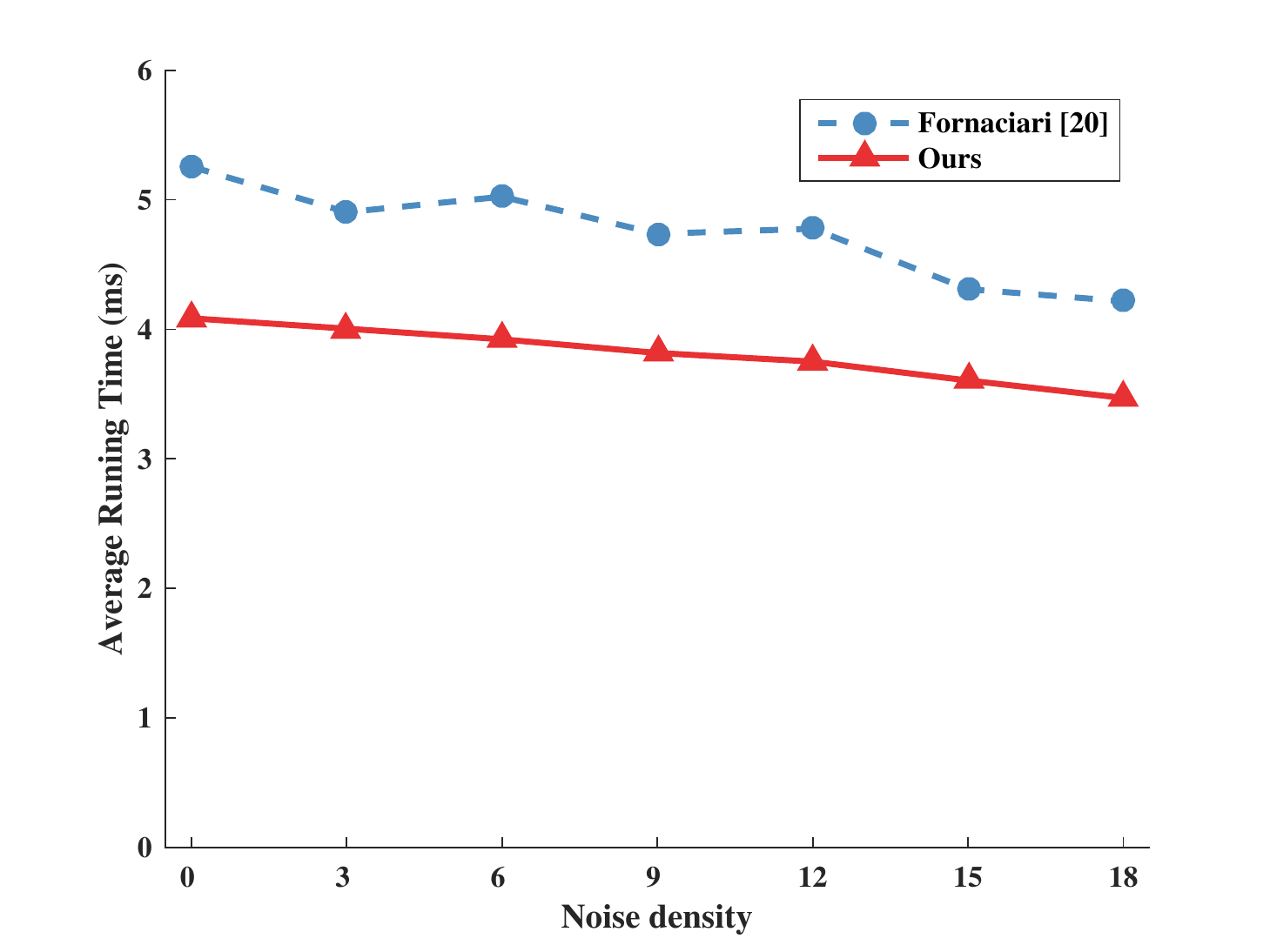} \label{fig:runningtime} }
    \caption{Robustness to salt-and-pepper noise. The ratios of noise to image are set as $3\%$, $6\%$, $9\%$, $12\%$, $15\%$ and $18\%$. (a) shows the F-measure of \cite{fornaciari2014fast} and our method. (b) Running time are plotted against noise density. Each point of these curves is obtained from the detection results of Dataset \#2.}
    \label{fig:salt}
\end{figure*}

We add salt-and-pepper noise to images of Dataset \#2 with the ratios of noise to image set as $3\%$, $6\%$, $9\%$, $12\%$, $15\%$ and $18\%$ to validate the robustness of our method to noise. Again, we compare with~\cite{fornaciari2014fast} on efficiency and accuracy, shown in~\fref{fig:salt}. The values of F-measure for these two methods on images without noise are $6.2$ and $5.8$, respectively. The values decrease for both methods when the noise increases. When the ratio of noise to image is up to 18\%, our value is about $0.35$, higher than that of~\cite{fornaciari2014fast}. When the noise level is getting higher, the execution time of our detector stays around $4$ms, and that of~\cite{fornaciari2014fast} fluctuates a bit around $5$ms not so stable as ours. Generally, both methods runs faster as noise increases because they are able to remove small arc segments caused by noise. As a summary, noise has few effect on the efficiency, but slightly lower down the accuracy.

\section{Conclusions}
In this paper, an ellipse detector for real-time application is proposed. We trades off accuracy and efficiency and pay more attention to the execution time. We introduce a new geometry constraint to prune lines and select arcs belong to the same ellipse. The detector removes the straight arcs based on characteristic on line  (CNL), and selects candidate elliptical arc combinations by characteristic on conic (CNC). Our method outperforms the-state-of-the-art by the experiments on real images, which can be used in real-time for various applications. In the future, we will improve our method to make it possible to detect quite small ellipses in images, which is a challenge for most existing methods.

\appendices
\section{Proof of Pascal's hexagon theorem}
\textbf{Pascal's hexagon theorem:} Let $\{{Q_i}^{(j)}|i=1,2,3;j=1,2\}$ be different points on a non-degenerative conic ${\cal C}$, as shown in~\fref{fig:ProveCNC}. Then three intersections
\begin{equation}
\left\{ \begin{gathered}
  {R_1} =  <Q_2^{(2)}Q_3^{(1)},Q_1^{(2)}Q_1^{(1)} > , \hfill \\
  {R_2} =  <Q_3^{(2)}Q_1^{(1)},Q_2^{(2)}Q_2^{(1)} > , \hfill \\
  {R_3} =  <Q_1^{(2)}Q_2^{(1)},Q_3^{(1)}Q_3^{(2)} >  \hfill \\
\end{gathered}  \right.
\end{equation}
are collinear.

\textbf{Proof:} The coordinate of $R_1$, $R_2$ and $R_3$ can be represented by $\{{Q_i}^{(j)}|i=1,2,3;j=1,2\}$ as
\begin{equation}
\left\{ \begin{gathered}
  {R_1} = ({Q_2}^{(2)}\times{Q_3}^{(1)})\times({Q_1}^{(2)}\times{Q_1}^{(1)}), \hfill \\
  {R_2} = ({Q_3}^{(2)}\times{Q_1}^{(1)})\times({Q_2}^{(2)}\times{Q_2}^{(1)}), \hfill \\
  {R_3} = ({Q_1}^{(2)}\times{Q_2}^{(1)})\times({Q_3}^{(1)}\times{Q_3}^{(2)}), \hfill \\
\end{gathered}  \right.
\end{equation}
where $\times$ denotes the cross product of two points. Then $R_1$, $R_2$ and $R_3$ can be represented through simple calculations as
\begin{equation}\label{eq:eqR3}
\left\{ \begin{array}{ll}
R_1 =& |{{Q_2}^{(2)}}, {{Q_3}^{(1)}}, {{Q_1}^{(1)}}|{{Q_1}^{(2)}} \\
&- |{{Q_2}^{(2)}}, {{Q_3}^{(1)}}, {{Q_1}^{(2)}}|{{Q_1}^{(1)}},\\
R_2 =& |{{Q_3}^{(2)}}, {{Q_1}^{(1)}}, {{Q_2}^{(1)}}|{{Q_2}^{(2)}} \\
&- |{{Q_3}^{(2)}},{{Q_1}^{(1)}}, {{Q_2}^{(2)}}|{{Q_2}^{(1)}},\\
R_3 =& |{{Q_1}^{(2)}}, {{Q_2}^{(1)}}, {{Q_3}^{(2)}}|{{Q_3}^{(1)}} \\
&- |{{Q_1}^{(2)}}, {{Q_2}^{(1)}}, {{Q_3}^{(1)}}|{{Q_3}^{(2)}}.
\end{array} \right.
\end{equation}
To prove that $R_1$, $R_2$ and $R_3$ are collinear is equivalent to prove
\begin{equation}\label{eq:sameline1}
  |R_1,R_2,R_3|=0.
\end{equation}
Then we can substitute~\eref{eq:eqR3} into~\eref{eq:sameline1}, and get the equivalent equation as
\begin{equation}
\begin{aligned}\label{eq:sixP}
&|{Q_1}^{(1)},{Q_2}^{(1)},{Q_3}^{(1)}||{Q_1}^{(2)},{Q_2}^{(2)},{Q_3}^{(1)}|\\
&|{Q_2}^{(2)},{Q_3}^{(2)},{Q_1}^{(1)}||{Q_3}^{(2)},{Q_1}^{(2)},{Q_2}^{(1)}| \\
&=|{Q_1}^{(2)},{Q_2}^{(2)},{Q_3}^{(2)}||{Q_1}^{(1)},{Q_2}^{(2)},{Q_3}^{(1)}|\\
&|{Q_2}^{(1)},{Q_3}^{(2)},{Q_1}^{(1)}||{Q_3}^{(1)},{Q_1}^{(2)},{Q_2}^{(1)}|.
\end{aligned}
\end{equation}
In order to prove it, we can replace any point ${Q_i}^{(j)}$ with the general point $Q(x,y,z)$, taking ${Q_1}^{(1)}$ for example, then we can get the parametric equations of conic $\cal C$ by other five points. As ${Q_1}^{(1)}$ is one point on conic $\cal C$, it must meet~\eref{eq:sixP}. The proof is completed.

\ifCLASSOPTIONcompsoc
  \section*{Acknowledgments}
\else
  \section*{Acknowledgment}
\fi

We thank Dr. M. Fornaciari, Dr. S. Mulleti and Dr. T. Lu for providing
their executables and insights. We also thank Dr. M. Fornaciari for
his experimental data.

\ifCLASSOPTIONcaptionsoff
  \newpage
\fi



\bibliographystyle{IEEEtran}
\bibliography{A_Fast_Ellipse_Detector_Using_Projective_Invariant_Pruning}
\end{document}